\PassOptionsToPackage{table}{xcolor} 
\documentclass[multiauthors,onecolumn]{LaTeXclass/cohere}
\usepackage{ragged2e}
\usepackage{lipsum}
\usepackage{pdfpages}
\usepackage{colortbl}
\usepackage{tabulary}
\usepackage{longtable}
\usepackage{array}
\usepackage{arydshln}
\usepackage{tipa}
\usepackage{placeins}
\usepackage{tablefootnote}
\usepackage{comment}
\usepackage{wrapfig}
\usepackage{breqn} 
\usepackage{booktabs}
\usepackage{multirow}

\usepackage{pifont}
\definecolor{deeppurple}{HTML}{9e02f7}
\definecolor{forestgreen}{HTML}{2e7d43}
\usepackage{multirow}
\usepackage{tabularx}
\usepackage{ragged2e} 
\newcolumntype{Y}{>{\RaggedRight\arraybackslash}X} 

\usepackage{cleveref}
\usepackage{wrapfig}
\usepackage{tcolorbox}
\usepackage{amsmath,amssymb,amsfonts,mathtools,amsthm}
\usepackage{mdframed}
\usepackage{fontspec}
\usepackage{fancyvrb}
\usepackage{fvextra}
\usepackage{float}
\usepackage{comment}
\usepackage{url}
\usepackage{booktabs}
\usepackage{microtype}

\newtcolorbox{promptbox}[1]{
    enhanced,
    breakable,
    colback=gray!5,
    colframe=gray!70,
    title={#1},
    fonttitle=\bfseries,
    boxrule=0.5pt,
    arc=2pt
}

\definecolor{lightgray}{gray}{0.45}

\definecolor{sectiongray}{gray}{0.2}
\definecolor{boxgray}{gray}{0.25}
\definecolor{lightgray}{gray}{0.65}

\definecolor{HeaderBG}{RGB}{35,35,35}   
\definecolor{HeaderFG}{RGB}{255,255,255}
\definecolor{FrameGray}{gray}{0.25}
\definecolor{DarkBlue}{HTML}{2D4CB9}

\title{The IOL-AI Challenge: 
}
\subtitle{An Open Challenge towards Advancing Linguistic Reasoning}

\author{name={Eduardo Sánchez\fa},affiliation={1}}
\author{name={Rita Berrada\fa},affiliation={2,3}}
\author{name={ Dan-Mircea Mirea\fa},affiliation={4}}
\author{name={Sara Rajaee}, affiliation={5}}
\author{name={Alexander Piperski}, affiliation={6}}
\author{name={Ana Meta Dolinar}, affiliation={5}}
\author{name={Boris Iomdin}, affiliation={7}}
\author{name={Andrey Nikulin}, affiliation={8}}
\author{name={Mariya Shmatova}, affiliation={9}}
\author{name={Marzieh Fadaee}, affiliation={3}}
\author{name={Julia Kreutzer\psa}, affiliation={3}}
\affiliations{
     \item[1] University College London \& Meta
     \item[2] CentraleSupélec \& McGill University
     \item[3] Cohere Labs
     \item[4] Princeton University
    \item[5] University of Amsterdam
     \item[6] Stockholm University
     \item[7] Faculty of Liberal Arts and Sciences
     \item[8] Universidade Federal de Goiás
     \item[9] Independent
 }

\abstract{

Reasoning in LLMs is overwhelmingly studied in domains that provide a model with rules: mathematics and code. Linguistic puzzles invert this: the solver must first discover the system before reasoning within it. We present the IOL-AI Challenge, an open-science competition run on the unseen problems of the  International Linguistics Olympiad (IOL) 2026 Individual Contest, evaluated both automatically and, for the first time, by members of the official IOL Jury under the same rubrics applied to human contestants. The challenge drew 731 submissions from 46 teams under a strict compute budget (one T4, 30 mins). We additionally benchmark 15 unconstrained frontier and open models, with Claude Opus 4.8 earning a jury score equivalent to a gold medal, while both resource-constrained systems we submitted for jury grading scored in the range of the bottom 5\% of contestants. Capability was not determined by scale: 14B submissions outperform models twice their size, and gains come from decoding and output-handling rather than model capacity. We also found that automatic metrics rank systems exactly as the jury does, but compress the scale, upscoring weak systems by ${\sim}13$ points and understating strong ones.
Our analysis shows that while frontier models might have prior knowledge about some of the problem languages, it does not significantly help them solve the linguistic reasoning tasks, leaving linguistic reasoning as a strong benchmarking proxy for generalizable reasoning skills. 
}
\begin{document}

\section{Introduction}

Reasoning has become one of the central paradigms for improving the capabilities of large language models~\citep{deepseekai2025deepseekr1incentivizingreasoningcapability,marjanovic2026deepseekr,qwen3.5,gemmateam2026gemma4}.
Rather than relying only on larger models, more data, or longer pretraining, recent systems increasingly improve their performance by allocating compute at inference time and generating intermediate reasoning~\citep{snell2025scaling,geiping2025scaling}, exploring alternative paths~\citep{qiu2026hyper,brown2024largelanguagemonkeysscaling}, verifying solutions, and revising their answers~\citep{hong-etal-2024-closer,zhang2025incentivizing,liu2025trust}.

The advances have been impressive yet the way we study reasoning has remained remarkably narrow.
The dominant tests of reasoning are mathematics~\citep{shen2025satori,chen2025rethinking,zhang2026realmath} and coding~\citep{chen2025dycodeeval,he2026evaluating}, which are domains with formal structures, unambiguous answers, and abundant automatically verifiable feedback.
However, real-world problems often do not begin with a well-specified system of rules.
Before solving the problem, a solver may first need to discover what the relevant structure is.
Most reasoning benchmarks abstract away this step: they provide the rules, the formalism, or the problem space, and evaluate how well a system can reason within it.
\textit{A more fundamental question is whether a system can infer a useful structure from evidence and then reason within the discovered structure.}
This ability is critical for generalization: Solving new problems in unfamiliar domains requires
identifying regularities, forming hypotheses, and testing them against new evidence.

Natural language offers a particularly compelling domain in which to explore this question.
Human language is an intricate system of compositional structure that is highly variable, irregular, and context-dependent.
Linguistic problems (sometimes referred to as \emph{linguistic puzzles})---a genre of puzzles where the solver is typically expected to disentangle linguistic phenomena of an unfamiliar language based on a limited corpus \citep{zaliznjak1963,derzhanskipayne-2010}---provide a natural testbed for this ability.
To understand a language that one has never encountered before, one must infer latent structure from sparse observations: identify recurring units, discover grammatical transformations, form hypotheses about meaning, and use these hypotheses to predict novel forms~\citep{aycock2025can}.

The majority of linguistic problems come from linguistic olympiads, the largest of which is the International Linguistics Olympiad (IOL)\footnote{\url{https://ioling.org/}}. This annual competition unites secondary school students from many countries (46 at the 2026 edition), who compete in solving linguistic problems. Each problem challenges contestants (1) to analyze linguistic data from one or more languages they have likely never seen before, (2) to deduce a set of rules that can fully explain the data, and (3) to use these rules to translate into and out of the unfamiliar language. Contestants' solutions are graded by an international body of linguists, polyglots and past IOL contestants called the IOL Jury, all of whom have substantial expertise in composing and evaluating linguistic problems.

Prior work has uniformly reported that even frontier reasoning models consistently underperform at solving such puzzles~\citep{garnham2026could,kocmi-etal-2025-findings-wmt25,lian-etal-2025-lobster}.
We take past benchmarking efforts three steps further:
(1) We engage the open-science community to advance linguistic reasoning of open and resource-constrained models in the form of a time-bound challenge with a leaderboard: \emph{The IOL-AI Challenge}.
(2) We partner with organizers of the IOL to benchmark models on completely \emph{new and unseen problems} from the IOL 2026 Individual Contest, which removes any concerns of benchmark leakage even for closed models.
(3) We obtain \emph{expert-level human evaluations} from the official IOL jury---which allows us to identify expert-level judgments of the state of linguistic reasoning, where previously we had to rely on shallow automatic matching with official solutions.

Our study thus constitutes the most recent, most thorough and in-depth evaluation of linguistic reasoning capabilities of LLMs at scale. In contrast to prior work, we find that proprietary models have caught up with the task, but open models lag wide behind, and under resource constraints mostly fail the task---indicating that generalizable reasoning is still a very open problem.

\section{Related Work}
\textbf{Reasoning Capabilities }
The evaluation of the reasoning capabilities of large reasoning models has recently become a principle of AI benchmarks to measure their progress \citep{qwen3.5,gemmateam2026gemma4}. Reasoning datasets cover a wide range of tasks including, but not limited to, mathematics and theorem proving \citep{aime25,wang2026polymath,yang2023leandojo}, code generation \citep{jimenez2024swebench,merrill2026terminalbench}, science and multidisciplinary \citep{rein2024gpqa, phan2025lastexam}, and visual puzzles \citep{chollet2026arcagi2newchallengefrontier}. Solving problems from the International Mathematical Olympiad (IMO)\footnote{\url{https://www.imo-official.org/}}, which is widely known as the most difficult math contest, has been a big goal in mathematical reasoning. In 2025, \citet{huangyang-2025-winninggoldatimo2025} showed how using a verification-and-refinement pipeline, which proved useful in mathematics given the easily verifiable correctness of the individual logic steps in the solution, drastically increases the performance of SotA modelson IMO problems and achieved a gold-medal score.

\textbf{Linguistic Reasoning Benchmarks}
\citet{sahin-etal-2020-puzzling} proposed PuzzLing Machines including Linguistic Olympiad examples. Later, \citet{bean2024lingoly} gathered a series of translation and linguistic reasoning problems from the UK Linguistic Olympiad, presenting the LingOly benchmark. Similarly, Linguini, IOLBENCH, and LINGOLY-TOO have been built using IOL problems, including a diverse set of low-resource languages \citep{sanchez2025linguini,goyal2025iolbenchbenchmarkingllmslinguistic,khouja2026lingolytoo}. While prior benchmarks focus only on correct final answers, LOBSTER \citep{lian-etal-2025-lobster} has provided step-by-step solutions for each problem using first an advanced LLM to generate them, and then human experts to post-edit solutions.

\textbf{Advances in Linguistic Reasoning with AI}
Advanced LLMs still struggle with linguistic reasoning, yet efforts to improve their performance remain surprisingly limited. Trying to fill this gap, \cite{zhu-etal-2025-evaluating} has shown that LLMs' performance improves in a multi-turn step-by-step setting. Moreover, providing analogical exemplars in the context enhances linguistic reasoning performance \citep{ramji-ramji-2025-inductive}.
\citet{garnham2026could} study test-time scaling of large-scale LLMs and find that while benchmark performance can increase by a few points, it remains well below math and commonsense reasoning benchmarks.

\section{Linguistic Reasoning at IOL}\label{sec:iol}

IOL started in 2003, growing steadily from just 36 contestants from 6 countries at its first edition in Borovets, Bulgaria to a record of 255 contestants from 46 countries at the current edition in Bucharest, Romania. Each participating country can send one or two teams of 4 contestants, usually selected through the country's national olympiad. National olympiads\footnote{\url{https://ioling.org/countries/}} vary in size and similarity to the IOL, some having a different format in terms of number and length of linguistics puzzles.
There are other local or international linguistic olympiads, such as the Online Olympiad in Linguistics.\footnote{\url{https://onling.org/}} The goals of IOL and other linguistic olympiads alike are: \emph{to popularize linguistics at the pre-university level, to} \emph{raise awareness of linguistic diversity and low-resource languages globally, and to promote problem-solving and linguistic skills beyond traditional school subjects}. For a detailed history of linguistic olympiads and a discussion of their declared goals, the reader is referred to \citet{martins-2022}.

\subsection{Linguistic Problems}
The concept of linguistic problems goes back to a workbook by \citet{gleason-1955} and was first formalized by \citet{zaliznjak1963}; more recent presentations of the genre include \citet{derzhanskipayne-2010,martins-2022,silvacosta-2022,nikulin-2024,neacsu-2024}. A core idea behind this concept is the \emph{principle of self-sufficiency}, that is, the problem must be solvable without prior knowledge of any foreign languages or advanced linguistic concepts. Low-resource languages are typically chosen in order to minimize the risk of any solver being familiar with the language, thus offering them unfair advantage. Solvers (student contestants in the case of IOL) are given enough information about a language, including but not limited to a few sentences and their translations, a numeral system, or a script sample. To solve a given problem, students need to infer the relevant grammatical, phonological, and lexical patterns from the data, and later apply it to answer the assignments. By default, contestants are expected to provide a theoretical explanation for their answers, using only concepts from secondary education curricula (such as \emph{consonant}, \emph{suffix}, or \emph{verb}). Contestants are not required nor expected to have received prior training in linguistics---even though in recent years there has been a surge in training materials, solving problems assigned in past installments of linguistics olympiads remains the most common way of preparing for IOL \citep{belikovetal-2006,derzhanskivelinov-2012,derzhanskivelinov-2013,radev-2013a,radev-2013b,silvacosta-2022,neacsu-2024}.

Linguistic problems are often classified by the subdiscipline of linguistics they illustrate---e.g., phonetics/phonology, morphology, syntax, lexical semantics, kinship terminology, number terms \citep{derzhanskiveneva-2018}, computational concepts \citep{littell-2013}, corpus linguistics \citep{iomdinetal-2013}, writing systems. They can also be classified by problem structure \citep{martins-2022,nikulin-2024,neacsu-2024}, the most common structures being ``Rosetta stone problems'' \citep{bozhanovderzhanski-2013}, which involve translation or transcription assignments based on a bilingual corpus; and ``chaos-and-order problems'', which involve matching assignments.

\begin{table*}[t]
\centering
\resizebox{\textwidth}{!}{%
\begin{tabular}{llp{6cm}p{6.3cm}}
\toprule
\textbf{Problem Structure} & \textbf{Prompt Type} & \textbf{Context} & \textbf{Question} \\
\midrule
Rosetta Stone & Translation & Here are some sentences in English and their Komnzo translations: ...& Translate into English in all possible ways: ...\\
Chaos-and-order& Fill-in Blanks & Here is the family tree of a family of Sakurabiat speakers. ... & Fill in the blanks (i–vi). There is only one possible answer for each blank, but each answer is not necessarily just one word. \\
Chaos-and-order& Match Letters & Here are some words and phrases in Yélî Dnye and their English translations in arbitrary order: ... & Determine the correct correspondences.\\
\bottomrule
\end{tabular}
}
\caption{Examples from the IOL 2026 Individual Contest, structured into context and question to prompt LLMs.}
\label{tab:question_types}
\end{table*}

\subsection{The IOL Problem Set} The IOL problem set consists of five problems meant to be solved individually by each contestant within 6 hours (the Individual Contest) and one problem designed to be solved collaboratively by national teams of 4 (the Team Contest). Any working language may be requested for the competition and assignments are equivalent in all working languages (\citealt{derzhanski-2013}). The problem set is crafted and multilingually rendered by the IOL Problem Committee in the year prior to each IOL.

The individual problem set at IOL 2026 included a problem on phonetics and orthography in Central Alaskan Yup'ik (\textbf{P1}); a chaos-and-order problem on lexical semantics, in particular semantics of color terms, in Yélî Dnye (\textbf{P2}); a Rosetta stone problem on syntax in Iquito (\textbf{P3}); a chaos-and-order problem on kinship terminology in Sakurabiat (\textbf{P4}); and a Rosetta stone problem on verb morphology in Komnzo (\textbf{P5}).

\subsection{Grading} Each problem is graded by a subset of the IOL jury. In the weeks leading to IOL, after the problem set has been assigned, each grading group develops a grading scheme for their problem, deciding on which criteria to grade and how to weight them. Grading schemes allocate points to assignments and the theory part (the explanation of the rules) separately, as criteria: each assignment is usually one criterion, whereas the theory portion contains multiple criteria, each corresponding to a main rule that the solvers need to identify.

At IOL, each contestant's submission for a problem is graded independently by at least two jurors within the grading group, with translation assistance from other jurors or external language specialists if needed. In the event of a discrepancy between two or more jurors regarding any single criterion, the discrepancy must be resolved through discussion until consensus is achieved.

\section{The IOL-AI Challenge 2026}
The challenge was set up as an open-science competition preceding the IOL (outreach efforts described in \Cref{app:accessibility}), with a compute-restricted testing environment and a truly unseen test set that no participants had access to. Submissions were accepted for a month until July 26, 2026 when the on-site IOL competition started.

\subsection{Technical Setup}

\textbf{The Task}
The dataset includes the data from the 5 problems of the English version of the IOL 2026 Individual Contest, a total of 14 sub-assignments or tasks that these problems collectively contain. We distinguish 3 types of tasks: (1) Translation, where given a context, it is asked to translate a sentence from ``Solverese'' (here: English) to an unseen extremely low-resource language, (2) Fill-in Blanks, where given a set of examples, the model is asked to fill the missing forms, and (3) Match Letters, where the task is to match a set of phrases to their correspondences. \Cref{tab:question_types} presents an example of test data with different types.
We split the tasks into a development (8 tasks) and a test set (6 tasks) balanced across task types, where the development set is used for computing automatic scores on a public leaderboard, while scores on the test set are not revealed during the competition, but only after its end, as part of a private leaderboard that contains scores for the complete set of tasks. One task is excluded from automatic scoring as it requires only an explanation that cannot be graded automatically.
Submissions eligible to participate in the human evaluation had to provide a short textual explanation with every task solution as well.

\textbf{Format} We follow Linguini in the task format~\citep{sanchez2025linguini} where every data point has a context containing the given problem and a query for the asked question.
Most problems require returning a list of answers, such as filling in a list of blanks, or an ordered list of letters.
Reference solutions may also contain multiple valid answers for each problem, if there are any ambiguities.
Any visual clues are turned into text, and diagrams into JSON format, following the setup in~\citep{kocmi-etal-2025-findings-wmt25}.

\textbf{Automatic Evaluation}
Following previous work on linguistic reasoning \citep{sanchez2025linguini,bean2024lingoly}, we employ ChrF~\citep{popovic-2015-chrf} and exact match (EM) to evaluate the final answers. We report the geometric mean (GM) of these metrics because ChrF will likely overestimate true answer quality (e.g., partial matches can be obtained from copying words from the task), and EM will likely underestimate it (e.g. answer formatting or minor morphological variations).
ChrF and EM are first computed on each individual example (taking the highest score across reference solutions if there are multiple), then averaged across examples.\footnote{At the time of the AI competition, points were not assigned to tasks yet, so while the final human evaluation is point-weighted (more difficult problems giving more points), the automatic scores are uniformly aggregated across problems.}

\textbf{Implementation}
The challenge runs on Hugging Face (HF), built on the open-source
\texttt{competitions} framework. A submission is a public HF model
repository holding a self-contained inference script alongside the model weights
themselves, that is run on an evaluation sandbox without network access and writes answers and optionally explanations into an output file.
This evaluation sandboxing is necessary to prevent leaking the IOL-AI data. The dataset was uploaded directly by members of the
IOL jury and problem committee (AP, DMM)
to a private HF dataset.
Neither participants nor other organizers ever saw the data.
Each submission runs as an isolated HF Job
on a single NVIDIA T4 GPU (16\,GB VRAM, 8 vCPUs, 30\,GB RAM) under a
30-minute wall-clock limit.

After each run, teams receive the score on the dev split (scored server-side based on the output file); failed submissions additionally return an excerpt from the error log (e.g., ``time limit exceeded'').
Teams have up to 10 submissions per day and select
two for the private leaderboard.

\subsection{Baselines}
We provided four basic baselines, submitted through the same
pipeline as participant entries. They relied on four different open-weights models of different scales, with slightly different amounts of prompt customization. \Cref{app:baselines} describes them in detail. The baselines were
visible to participants on the leaderboard throughout the competition.
\Cref{tab:baselines} reports their automatic scores.

\begin{table}[h]
\centering
\small
\begin{tabular}{lccc}
\toprule
\textbf{Baseline} & \textbf{chrF} & \textbf{EM} & \textbf{GM} \\
\midrule
Qwen2.5-14B-Instruct-AWQ        & 19.70 & 4.49 & 9.40 \\
Tiny Aya Global (3.35B)                & 16.09 & 4.10 & 8.12 \\
Qwen2.5-7B-Instruct (NF4)       & 19.65 & 2.56 & 7.10 \\
Qwen2.5-1.5B-Instruct (starter) & 12.51 & 1.28 & 4.00 \\
\bottomrule
\end{tabular}
\caption{\textbf{Automatic evaluation scores of organizer baselines}. EM = exact match, GM = geometric mean.}
\label{tab:baselines}
\end{table}

\subsection{Extended Model Benchmarking}
To complement the resource-constrained submissions, we also benchmark 15 existing reasoning models (details in \Cref{app:extended_models}), including ``frontier'' models (open and proprietary), i.e., the largest, most resource-consuming, most recent, and reasoning-heavy models, and open ``mid-size'' models to bridge the gap to the IOL resource constraints.
Given that the latter ones are individually not competitive with the former, we aggregate the answers of four mid-sized models (Gemma 4 31B, Qwen 3.6 27b, DeepSeek R1 32B, GLM 4.7 Flash) in a best-of-$n$ ensemble (BoN; oracle selection: choosing the best scoring model for each task) to measure the ceiling of what these models can currently achieve, inspired by earlier ensemble success~\citep{garnham2026could}.

\subsection{Human Evaluation at the IOL}

We consider the top 10 challenge submissions with provided explanations along with the best closed- and open-source reasoning models for human evaluation, before selecting five systems to undergo human evaluation (see \Cref{sec:human}). 
Human evaluation is performed by a subset of the IOL jury (at least 2 graders per problem) using a very similar procedure to the grading of contestant solutions at IOL, including anonymization with respect to the model information and resolution of discrepancies between jury members through discussion. Details are described in \Cref{app:jury}.

\section{Results}\label{sec:results}
We first discuss the results from the automatic evaluation stage (\Cref{sec:automatic}), covering a larger set of models, before zooming in on the five selected submissions for the human evaluation (\Cref{sec:human}). Lastly we perform an analysis on the knowledge leveraged by top-scoring models (\Cref{sec:analysis}).

\subsection{Automatic Evaluation}\label{sec:automatic}

\subsubsection{Submissions from the Open Challenge}

We received 731 submissions (80.7\% ran successfully) from 46 external teams over the one-month
window. We describe submission statistics in more detail in \Cref{app:submissions}, and focus on the top 10 here, listed in \Cref{tab:top10}. Nine of the top
ten entries run a quantized 14B Qwen model, the largest model to comply with the compute constraint.
The best submission more than doubles our strongest baseline
with the same model (GM 9.40 $\rightarrow$ 19.79), showing that decoding and output handling are effective levers under constrained model size.

\label{sec:leading-submissions}
\begin{table}[H]
\centering
\small
\begin{tabular}{@{}r l c c c l@{}}
\toprule
\textbf{\#} & \textbf{Team} & \textbf{chrF} & \textbf{EM} & \textbf{GM} & \textbf{Model} \\
\midrule
1 & arvindcr4$^{\star}$ & 31.57 & 12.41 & 19.79 & Qwen2.5-14B AWQ \\
2 & BigRatz$^{\star}$ & 30.75 & 11.55 & 18.85 & Qwen2.5-14B AWQ \\
\hdashline
3 & hhhar & 29.14 & 11.35 & 18.19 & Qwen2.5-14B AWQ \\
4 & friedspaghetti & 28.98 & 11.07 & 17.91 & Qwen2.5-14B AWQ \\
\hdashline
5 & mastermind & 27.78 & 10.07 & 16.73 & Qwen2.5-14B AWQ \\
6 & Hul & 28.41 & 9.84 & 16.72 & Qwen2.5-14B AWQ \\
7 & jbuaba & 27.68 & 9.84 & 16.51 & Qwen2.5-14B AWQ \\
8 & attractordynamics & 27.56 & 9.12 & 15.85 & Qwen2.5-14B AWQ \\
\hdashline
9 & LOPE-NTU & 23.78 & 9.62 & 15.12 & Qwen3.5-4B Q8 GGUF \\
10 & srikarkashyap & 26.99 & 7.98 & 14.67 & Qwen3-14B AWQ \\
\bottomrule
\end{tabular}
\caption{\textbf{Automatic evaluation scores for the top 10 from the leaderboard of the IOL-AI challenge.}
Submissions with $\star$ were selected for jury evaluation. Dashed lines indicate first, second, and third place winners on the leaderboard.}
\label{tab:top10}
\end{table}

\textbf{Technique adoption and measured effects }
We inspected the archived repositories
and
tagged the techniques used in each submission. Where a team submitted
runs both with and without a technique, we measure its effect as the
within-team difference in mean private score ($n$ = teams with runs of
both kinds), this controls for stronger teams simply trying more
things. Greedy decoding was near-universal (38 teams), and 21 teams
additionally used temperature sampling, mostly for retries.
Self-consistency voting (24 teams) adds $+2.08$ points ($n=21$). Retry
or reformat loops (17 teams) add $+1.40$ ($n=14$). Chain-of-thought
prompting was widely tried (33 teams) but costs $-1.55$ ($n=25$) and
none of the top ten submissions used it. JSON-structured output (16 teams) is the most
damaging at $-4.66$ ($n=10$). Fine-tuning or LoRA on past puzzles
brought no gain, and neither did retrieval, consistent with problems
that are self-contained by construction.

\textbf{Leading Approaches }
Leading submissions converge on a common recipe. They lower the
repetition penalty from its default greedy-mode value of 1.05 to 1.0, a
silent parameter that penalizes generating the same characters several
times in a row, a frequent pattern in these languages. None uses
chain-of-thought prompting: it was tried repeatedly, and runs without it
consistently scored higher. All top entries anchor on a greedy pass, and
when they sample, draw retries at temperature 0.5 as time allows,
replacing the greedy answer only on agreement. Token budgets are set
adaptively from the wall-clock time remaining, and the submission file
is written incrementally. Part of this convergence reflects copying,
since submission repositories were public throughout the competition:
the 1st and 5th entries run the same script, differing only in
whitespace, and their score difference reflects run-to-run variance
rather than any difference in method. We return to this in the
Limitations. The individual pipelines are described in
\Cref{app:submissions}.

\subsubsection{Extended Models}

\begin{table}[ht]
\centering
\begin{tabular}{lcccc}
\toprule
\textbf{Model} & \textbf{chrF} & \textbf{EM} & \textbf{GM} & \textbf{LFMR} \\
\midrule
\multicolumn{5}{l}{\textbf{Proprietary Models}} \\
\midrule
Claude Opus 4.8 $\star$ & 82.9 & 68.1 & 75.1 & 100.0 \\ 
 Gemini 3.6 Flash $\star$ & 70.6 & 45.0 & 56.4 & 100.0 \\
 GPT 5.6 Sol  & 63.3 & 43.1 & 52.2 & 100.0 \\ 
\midrule
\multicolumn{5}{l}{\textbf{Open Large Models}} \\
\midrule
Kimi K2.6 & 57.1 & 35.8 & 45.2 & 100.0 \\ 
 GLM 5.2 & 58.0 & 35.0 & 45.0 & 92.9 \\
 DeepSeek V4 Pro & 53.6 & 31.5 & 41.0 & 92.9 \\ 
 Qwen 3.7 Max & 52.2 & 29.4 & 39.2 & 100.0 \\ 
Inkling & 48.0 & 24.3 & 34.1 & 100.0 \\ 
 Nemotron 3 Ultra & 41.5 & 20.1 & 28.9 & 64.3 \\ 
  Command A+ & 34.1 & 11.1 & 19.4 & 100.0 \\
 MiniMax M3 & 14.4 & 7.9 & 10.6 & 78.6 \\
 \midrule
\multicolumn{5}{l}{\textbf{Open Mid-Size Models}} \\
  \midrule
  Qwen 3.8 27B (think)    &        51.2 &  30.9  & 39.8  & 92.9 \\
  Glimmer 30B (high)    &    48.7 &  30.3  & 38.4 &  85.7 \\
  Qwen 3.6 27B (think)    &        37.1 &  19.2  & 26.7  & 92.9 \\
   Gemma 4 31B   & 37.0 & 14.6 & 23.2 &  92.9 \\
   Qwen 3.6 27B     & 25.2 & 14.9 & 19.4 &  64.3 \\
   DeepSeek R1 32B & 23.7 &  3.0 &  8.5 & 100.0 \\
   GLM 4.7 Flash   &  6.6 &  0.0 &  0.0 &  21.4 \\
  \hdashline
   BoN Open (oracle) $\star$ & 48.5 & 23.2 & 33.5 & 100.0 \\
\bottomrule
\end{tabular}

\caption{\textbf{Automatic evaluation scores for the extended model set on the IOL 2026 problems} (after custom answer parsing).
LFMR = line format match rate. Model outputs that were chosen for the jury are marked with $\star$.}
\label{tab:results_extended}
\end{table}

\begin{figure}[t]
  \centering
  \includegraphics[width=0.9\columnwidth]{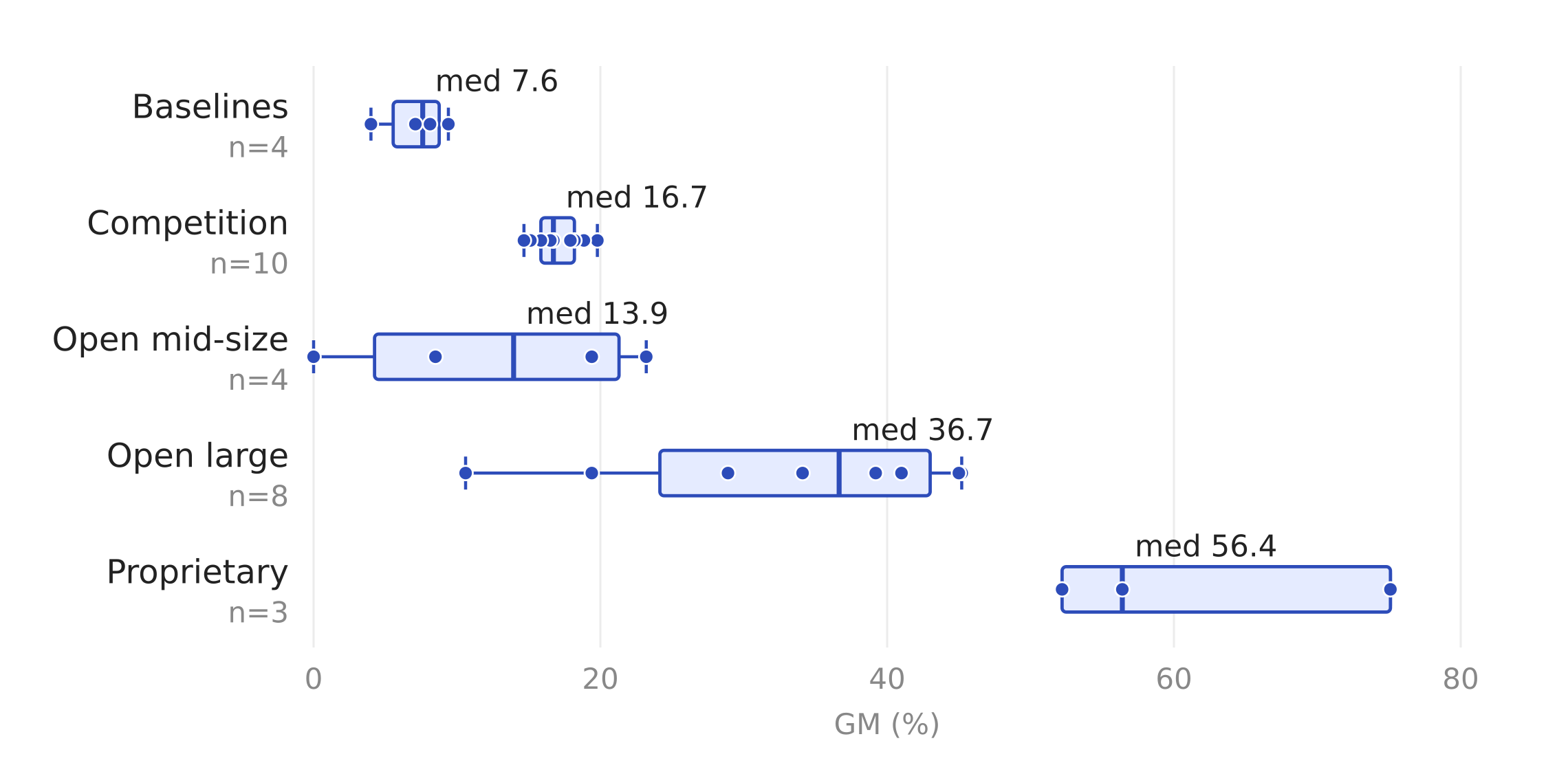}
  \caption{Distribution of automatic scores (GM) by system tier, spanning from organizer baselines and resource-constrained competition submissions to open mid-size, open large, and proprietary frontier models.}
  \label{fig:gm_by_tier}
\end{figure}

\Cref{tab:results_extended} contains the automatic evaluation scores for our
selected 15 individual models and the ensemble. We report line format match rate (LFMR) besides the quality scores to indicate where low scores are due to low format compliance. Note that all submissions undergo a custom answer parsing process that e.g. strips markdown (\Cref{app:extended_models}).
As expected from evaluations in prior work, proprietary models perform strongest, with a noticeable 19-point gap between Claude-Opus-4.8 and the following Gemini-3.6-Flash and GPT5.6-Sol. Open (large) models follow after a 7 point gap, with a large range in scores, going down to 10.6. Of the mid-sized models, Gemma4-31B-It scores highest (23.2), and the ensemble gains another 10 points on top of it, catching up with some of the larger open models. Performance is therefore not strictly determined by scale: as \Cref{fig:gm_by_tier} shows, resource-constrained submissions at 14B double the organizer baselines and match or exceed the unconstrained open mid-size tier, though all open systems remain far below the frontier. As linguistic reasoning is not a typical post-training task, and most models struggle with the output format, ``tweaking'' inference goes a long way.

\subsubsection{Comparison across Tasks}

\begin{figure*}[p]
  \centering
  \begin{subfigure}[t]{\textwidth}
    \centering
    \includegraphics[clip=true, trim=0.2cm 1.1cm 0cm 0.72cm,width=\textwidth]{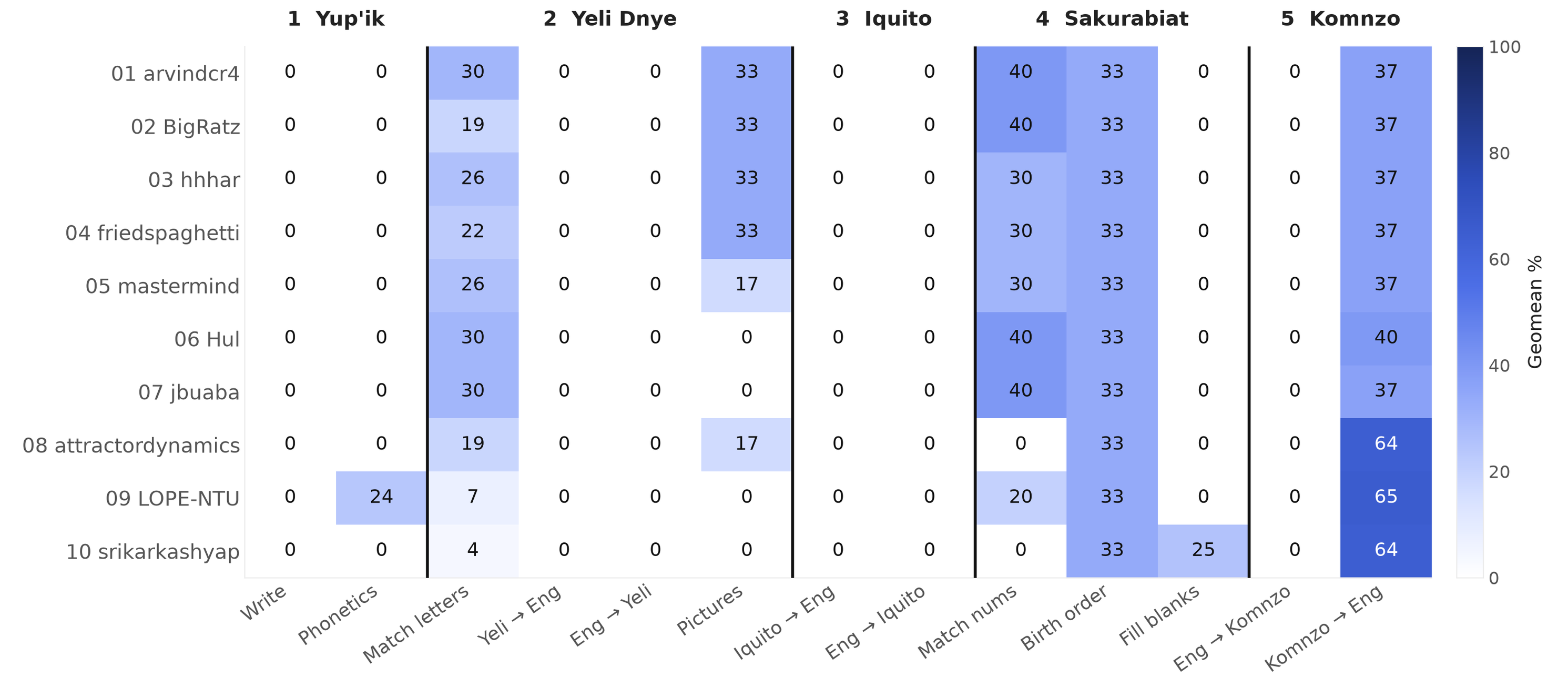}
    \caption{Top 10 submissions from the IOL-AI challenge.}
    \label{fig:heatmap-submissions}
  \end{subfigure}
  \vspace{1em}
  \begin{subfigure}[t]{\textwidth}
    \centering
    \includegraphics[clip=true, trim=0.2cm 1.1cm 0cm 0.72cm,width=\textwidth]{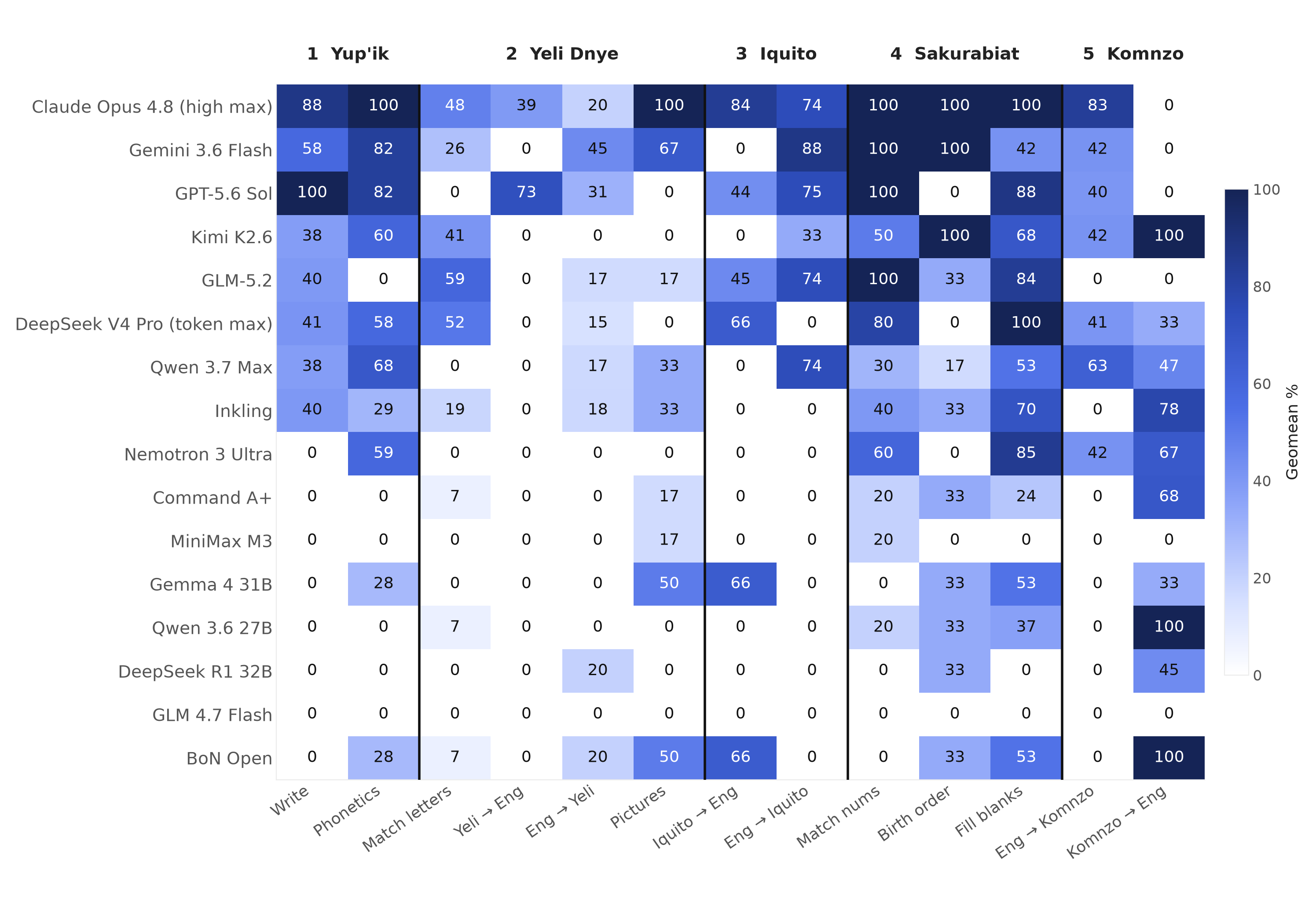}
    \caption{Frontier models.}
    \label{fig:heatmap-frontier}
  \end{subfigure}
  \caption{Breakdown of scores (geomean \%) across the individual tasks of the 5 IOL 2026 problems, for the top 10 challenge submissions (top) and frontier models (bottom).}
  \label{fig:heatmaps}
\end{figure*}

Among the top 10 submissions (\Cref{fig:heatmap-submissions}), difficulty separates cleanly by problem.
Iquito (\textbf{P3}) is the hardest: both of its tasks score 0\% exact match for all ten teams.
Yup'ik (\textbf{P1}) and Y\'el\^i~Dnye (\textbf{P2}) follow closely, with no team solving Yup'ik writing or either Y\'el\^i~Dnye translation direction, and only the closed-form items (Y\'el\^i~Dnye letter and picture matching, Yup'ik letter matching) earning partial credit.
On these unsolved tasks, chrF often remains in the 23 to 53 range, indicating that answers contain some correct components.
Submissions perform best on Sakurabiat (\textbf{P4}), whose short, closed items (birth order, number matching) are answered by nearly every team.
Komnzo (\textbf{P5}) shows the largest asymmetry within a single problem: Komnzo$\to$English is the best-solved task across all submissions, while English$\to$Komnzo receives 0\% throughout.
Teams are thus able to analyze Komnzo forms well enough to translate them into English, but not to produce correct forms in the language itself.

Frontier models (\Cref{fig:heatmap-frontier}) score higher on every problem but rank them differently.
Y\'el\^i~Dnye (\textbf{P2}) becomes the hardest problem (about 20\% geomean), with both translation directions near 11\%, whereas Iquito (\textbf{P3}), unsolved by all submissions, moves to the middle of the range.
Sakurabiat (\textbf{P4}) remains the easiest (about 55\%), and the Komnzo (\textbf{P5}) asymmetry persists: English$\to$Komnzo stays low (about 28.5\%) even for models that succeed in the opposite direction.
On most problems, scores follow overall model strength, but translation tasks diverge from this pattern: only proprietary models obtain non-zero scores on Y\'el\^i~Dnye$\to$English, while the same models fail Komnzo$\to$English, where several open models earn substantial credit.

\FloatBarrier
\subsection{Human Evaluation by the IOL Jury}\label{sec:human}

We selected the top 2 frontier models according to their automatic evaluation scores (Claude Opus 4.8 and Gemini-3.6-Flash) for human evaluation. From the top submissions we selected the first one, and then manually went through subsequent ranks and verified the explanations, and found that many of them were either templated or meaningless (copies of the same sentence). The submission ranked 11th stood out as it provided structured explanations with schemata, so we selected it to provide a contrastive solution.
From the open mid-size models, we submitted the ensemble, representing the ceiling of what a team of mid-size models can achieve if assembled right.
In this way, we hope to capture the full spectrum from heavily tuned, severely resource-constrained systems, over an ensemble of mid-size open models, to proprietary frontier models.

\begin{table*}[ht]
    \centering
    \resizebox{\textwidth}{!}{%
    \begin{tabular}{l|cc|cccccccccc}
    \toprule
    & \multicolumn{2}{c|}{Total Score} & \multicolumn{2}{c}{P1} & \multicolumn{2}{c}{P2} &\multicolumn{2}{c}{P
    3} &\multicolumn{2}{c}{P4} &\multicolumn{2}{c}{P5}\\
    Model Name & Jury & Aut. & Jury & Aut. & Jury & Aut. & Jury & Aut. & Jury & Aut. & Jury & Aut.\\
    \midrule
        Claude Opus 4.8 & 79.5 & 75.1 & 18.5    & 14.5 & 10.5 & 16.0 &    15.5 &12.2 & 19 & 23.2&    16 & 9.2\\ 
        Gemini 3.6 Flash & 60.5 & 56.4 & 13& 10.8&    9    &11.5 &15& 9.2 &    13& 19.0&    10.5 & 5.8\\ 
        BoN Open (oracle) & 20.3 & 33.5 & 4.5 & 3.4 & 2.3 & 6.7 & 4.5 & 7.2 & 3.5 & 7.3 & 5.5 & 8.9 \\ 
        arvindcr4 (\#1) & 5.9 & 19.8 & 0.8 & 1.3 & 2.1 & 5.8 & 0.5 & 2.6 & 2 & 6.3 & 0.5 & 3.8 \\ 
        cabanosss (\#11) & 2.5 & 14.6 & 1 & 1.0 & 1 & 5.3 & 0 & 2.2 & 0 & 2.3 & 0.5 & 3.8 \\ 
    \bottomrule
    \end{tabular}%
    }
    \caption{\textit{Jury} evaluation results by problem for the selected AI submissions compared to the automatic GM scores (\textit{Aut.}).}
    \label{tab:jury_scores_problem}
\end{table*}

\subsubsection{Quantitative Comparison} 
\begin{figure*}[!htb]
    \begin{subfigure}[b]{0.48\textwidth}
        \centering
        \includegraphics[width=\linewidth]{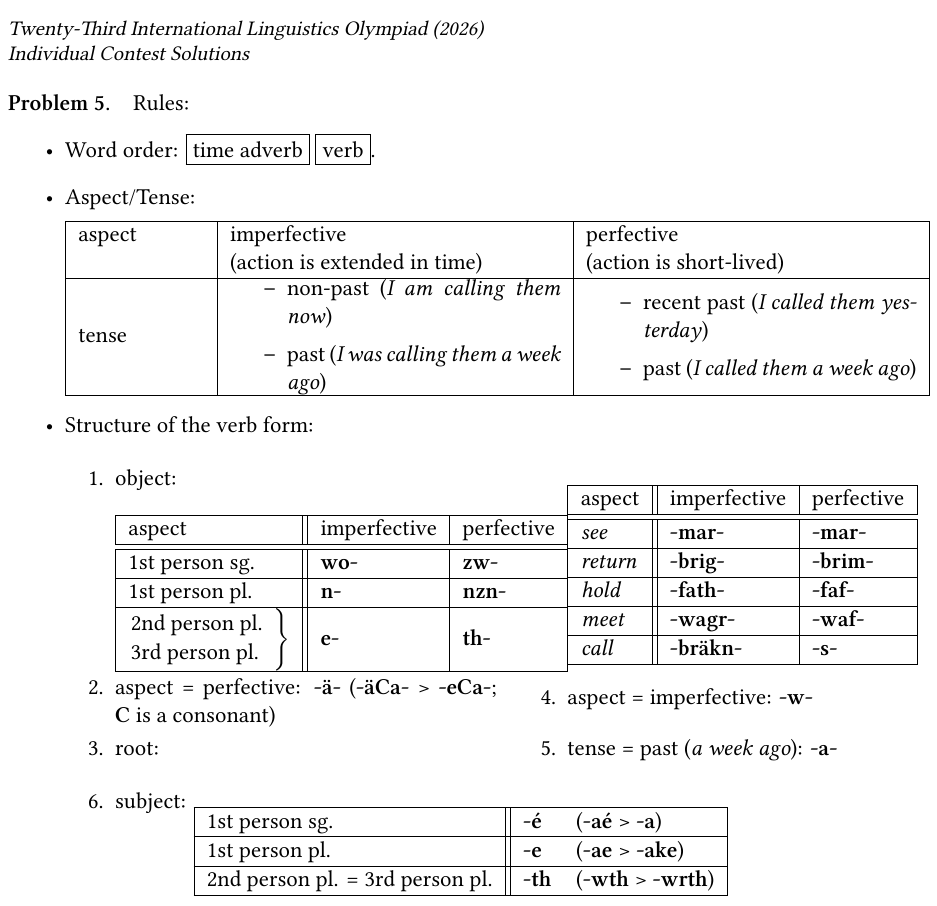}
        \caption{IOL solution explanation.}
        \label{fig:gold-explanation}
    \end{subfigure}
    \hfill
    \begin{subfigure}[b]{0.48\textwidth}
        \centering
        \includegraphics[width=\linewidth, trim={1.5cm 6cm 1.5cm 3cm}, clip]{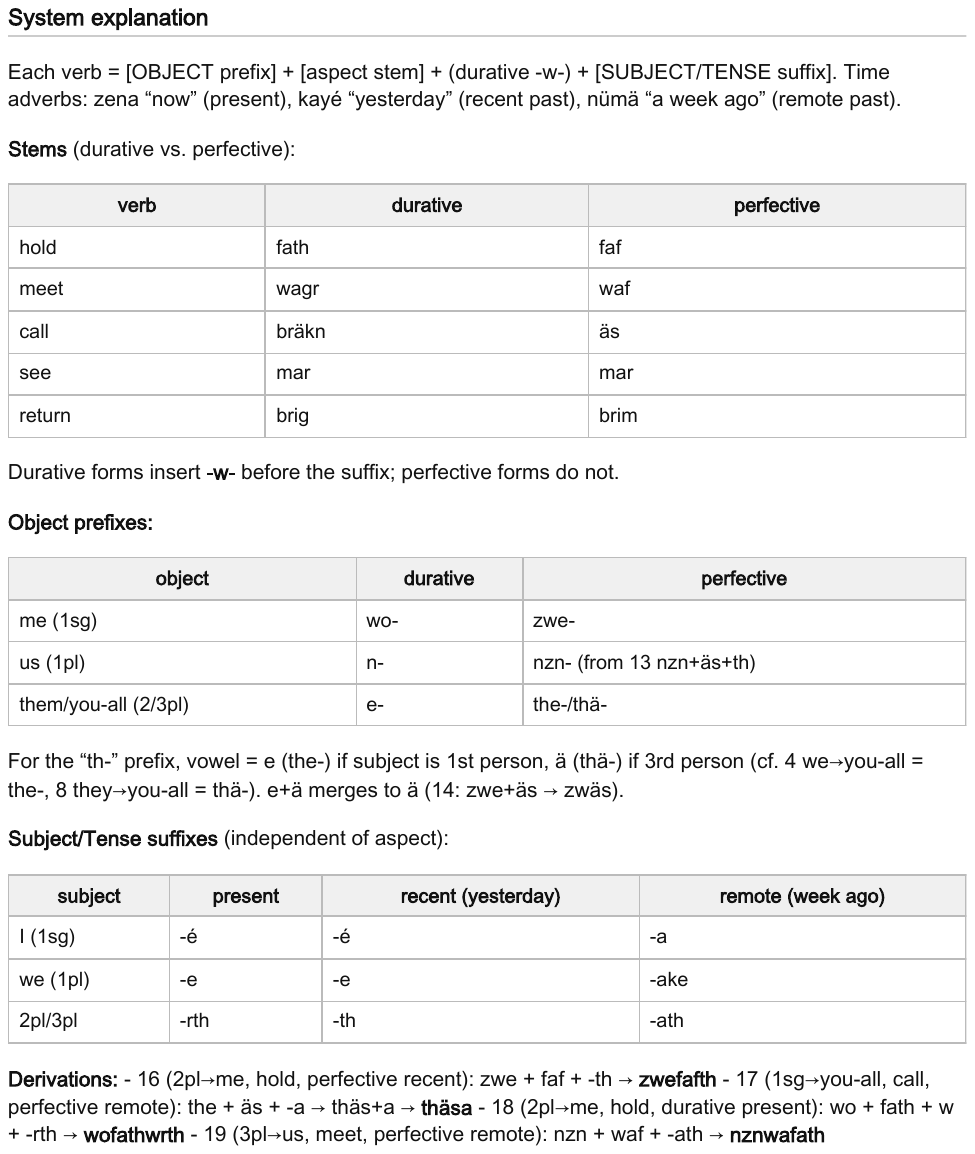}
        \caption{Claude 4.8 Opus explanation.}
        \label{fig:gpt-explanation}
    \end{subfigure}
    \caption{Schematic explanations provided in official IOL solutions for P5 in comparison to the top scoring model's solution explanation that we elicit through prompting.}
    \label{fig:explanation}
\end{figure*}

\begin{figure*}[!htb]
  \centering
  \includegraphics[width=\textwidth]{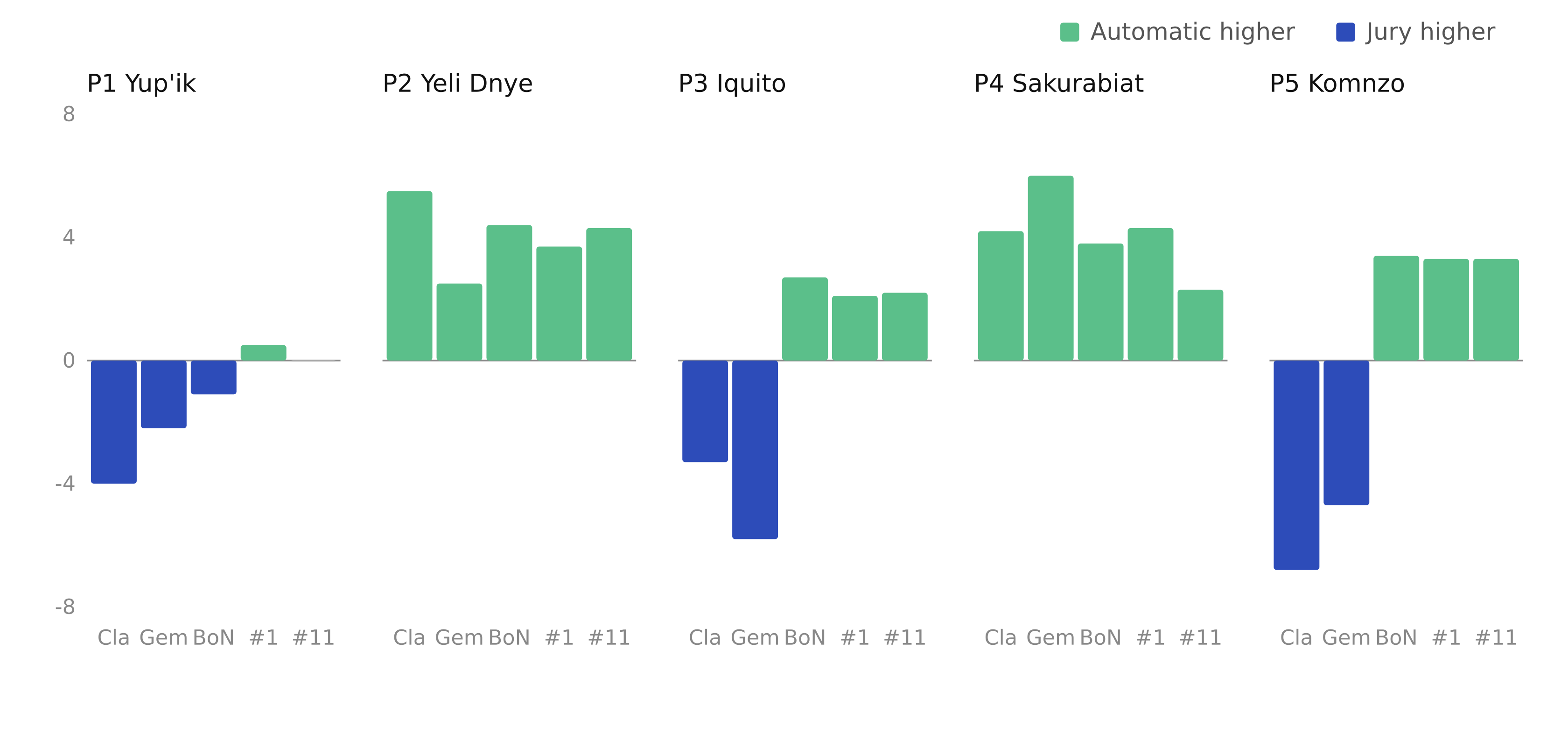}
  \caption{Difference between automatic and jury scores per problem for the five systems selected for human evaluation. Cla = Claude Opus 4.8, Gem = Gemini 3.6 Flash, BoN = BoN Open (oracle), \#1 = arvindcr4, \#11 = cabanosss.}
  \label{fig:aut_vs_jury}
\end{figure*}

The scores given by the IOL jurors showed the same rank-order of the models as the automatic score (\Cref{tab:jury_scores_problem}). Notably, Opus 4.8 received a score equivalent to that of an IOL gold medalist (lowest gold medal this year: 70.0), a score that would have placed it in fourth place in this year's ranking. 
An example of its explanations for P5 are shown side-by-side with the reference solution in \Cref{fig:explanation}.
Gemini 3.6 received a score equivalent to that of a silver medal (lowest silver medal this year: 58.4). The rest of the models were below the threshold for honorable mentions (35.3), meaning they all performed worse than 50\% of contestants this year. In fact, the two models from the challenge would have both ranked in the bottom 5\% of scores.
Agreement between the two protocols is near-perfect in rank (Spearman $\rho=1.00$, Pearson $r=0.99$, $n=5$) but not in magnitude. The two proprietary models score \emph{higher} with the jury than automatically ($+4.4$ and $+4.1$ points), whereas the ensemble and both challenge submissions score substantially \emph{lower} ($-13.2$, $-13.9$ and $-12.1$); the jury's range therefore spans 77.0 points against 60.5 automatically. \Cref{fig:aut_vs_jury} breaks these differences down by problem: the automatic metric sits above the jury on nearly every problem for the weaker systems, while the direction flips for the two proprietary models on P1, P3 and P5. Automatic scores overestimate the quality of the weaker submissions, due to absence of explanation scoring and overestimating influence of ChrF in the geometric mean; conversely, strong systems are under-credited, because a correct analysis stated in prose earns jury points but nothing automatically.

\subsubsection{Qualitative Comparison}
Beyond the quantitative differences, the IOL jurors reported several recurring qualitative differences between the AI solutions and the human solutions they usually grade: 
\begin{enumerate}[leftmargin=*,itemsep=1pt]
    \item \textbf{Local rather than consistent explanations.} Because our prompting procedure elicited a separate explanation for each task within a problem, the models sometimes gave different, and occasionally contradictory, analyses of the same phenomenon. 
    Strikingly, in P2--Yélî Dnye, the Gemini model proposed an incorrect theory in which nouns for objects were formed by reduplication from color terms for earlier assignments. However, for assignment 4 (which was unusual in that it explicitly asked for an explanation of how color terms are derived), the model fully switched its explanation to the canonical one (that some color terms are formed by reduplication while others are formed by comparison).
    This suggests that the model was in principle able to recover the right set of rules, but that it did not get to this explanations in the other assignments.

    \item \textbf{Verbosity: explanation traces and examples.} The models were generally more verbose than human contestants. They often gave entire explanation traces, something that contestants are explicitly instructed not to do, and which they can learn not to do from seeing canonical solutions. The models also often tried to explain a rule by giving an example sentence instead of the specific rule application (e.g.~in P3--Iquito, giving an example verb in the past tense and an example verb in the future tense, rather than stating the suffixes used to mark each tense), which is likewise not rewarded with any marks.

    \item \textbf{Hallucinations.} Lower-performing models sometimes hallucinated data (e.g., new terms such as \textit{completely black} in P2--Yélî Dnye, based on \textit{black} and \textit{completely white} which appear in the problem; or sounds that do not appear in the problem, such as [\textepsilon] or syllable-initial consonant clusters in P1--Yup'ik), outputted impossible answers to assignments (e.g., wrong number of syllables due to miscounting in P1), or provided non-sensical explanations and examples (e.g., \textit{arvindcr4} writing in P5-Komnzo \textit{for example, “thafa” (hold) with “th” suffix indicates the subject, and “wath” indicates the object}).

    \item \textbf{Underspecified rules.} Lower-performing models tended to hedge in their explanations using words such as \textit{typically} or \textit{often} (e.g., \textit{arvindcr4} in P3--Iquito: \textit{The verb typically comes after the subject}, which is always true and hence \textit{typically} is not needed).
    At IOL, such qualifications matter: a rule is expected to state precisely what the data support.

 \item \textbf{Undergeneralization.} The frontier models tended to miss broad patterns that human contestants sometimes correctly delineated. For instance in P5--Komnzo, no model parsed out morpheme \textit{-a-} appearing in sentences in non-recent past.
    This suggests AI models could have difficulty generalizing from sparse data during linguistic reasoning.

    \item \textbf{Overfitting or convoluted explanations.} 
    Explanations were often more complex than those of human contestants (e.g., in P5--Komnzo, Gemini incorrectly parsed the \textit{-s-} stem from \textit{nznästh} and \textit{zwäsath} and added a phonological modification (\textit{The punctual stem for “call” -näs- reduces to -äs- when preceded by the labial glide /w/ in the 1s punctual prefix zw- (e.g., zw-näs-ath → zwäsath).})

   \item \textbf{Technical terminology without a complete analysis.} The frontier models (Claude and Gemini) sometimes used technical linguistic terminology (e.g., \textit{iambic feet} in P1--Yup'ik, \textit{durative aspect} in P5--Komnzo, or a specific kinship-system annotation (\textit{MB} for mother's brother and \textit{FZ} for father's sister in P4--Sakurabiat). However, these terms were not always used consistently or completely. (e.g., writing \textit{MB's son} instead of the usual \textit{MBS} in kinship terminology).
\end{enumerate}

Overall, we found that models do not fail to identify linguistic structure altogether---the strongest models clearly are able to achieve medal-level scores. However, there are differences in how that structure is used across a problem: human solutions tend to build a compact analysis that explains the data as a whole, while model solutions more often reason from one assignment to the next. This difference is easy to miss when evaluation considers only the final answer, but becomes much more visible when the theory explanations themselves are graded.

\begin{figure}[t]
  \centering
  \includegraphics[width=0.7\columnwidth]{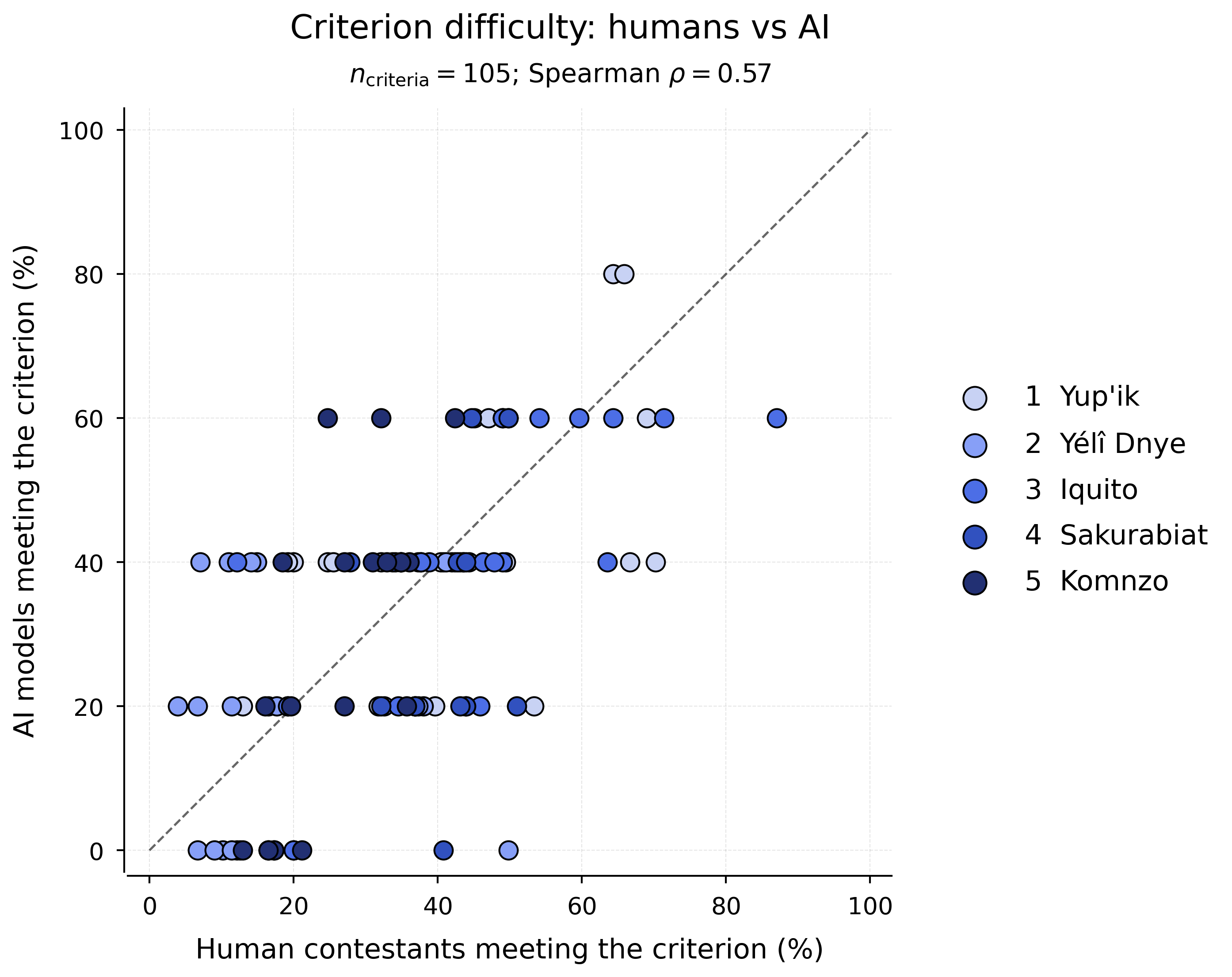}
  \caption{Comparison of the difficulty of criteria in the official IOL grading schemes between human contestants and models. Difficulty is measured as the percentage of solutions that meet the criterion and receive the respective mark, split into human contestants and AI models.}
  \label{fig:criterion_difficulty}
\end{figure}

On top of the differences, the jury also highlighted similarities to human contestants' solutions, such as the order in which phenomena are described in the explanations (e.g., tending to start with word order in P5--Komnzo), or the rank-order of difficulty of the different phenomena. Indeed, in line with the jurors' impressions, the difficulty level of all theory and explanation criteria across the 5 grading schemes (measured as the \% models or contestants that met the criterion, i.e., solved that subunit of the problem correctly) correlated strongly between humans and models (Spearman $\rho = 0.57$; \Cref{fig:criterion_difficulty}). This link was observed despite the fact that some problems seemed overall easier for models (P5--Komnzo), whereas others were easier for human contestants (P3--Iquito). This finding suggests that linguistic reasoning comes in different levels of difficulty that is to some extend intrinsic, irrespective of whether the solver is human or machine.

\subsection{Analysis: Knowledge Probing}\label{sec:analysis}
A key question is which knowledge successful models rely on (not only on reasoning capabilities)---one may speculate whether they post-trained on linguistic tasks, or have included grammar books in their pretraining more than others.
We cannot perform training data analysis for most benchmarked models due to missing data releases, but we can estimate proxies of their leverage of prior knowledge by relating the following estimates to problem solving success:
\begin{enumerate}
    \item Measure availability of \textit{online resources} in or about the problem languages (web crawl statistics, Wikipedia article length, number of Glottolog~\citep{glottolog} listed resources). This estimates how much the models could have had prior exposure with the language or words in that language.
    \item Probe models about their \textit{explicit knowledge about the languages} by turning the contextual language facts of the problem context in a multiple-choice task, and the lexical items from the problem language into a language identification task (also multiple-choice). This gives us proxies about (a) prior language knowledge and (b) lexical familiarity or ability to deduce language identity from single words.
    \item \textit{Remove the language name and background information} from the problem context and measure how scores change across problem languages. This tell us which models are able to leverage this information to bias their predictions, and how much they rely on it for solving the problems.
\end{enumerate}

\textbf{Presence on the web } All problem languages would fall under the category of (extremely) ``low-resourced'' languages in NLP~\citep{joshi-etal-2020-state} beyond the scope of mainstream AI technology development~\citep{blasi-etal-2022-systematic}. However, Yup'ik (\texttt{esu}) and Yélî Dnye (\texttt{yle}) documents can be found in FineWeb-2~\citep{penedo2025fineweb2pipelinescale} and GlotCC-V1~\citep{kargaran2025glotccopenbroadcoveragecommoncrawl} web crawls,\footnote{According to iso codes; without native speaker confirmation that these documents are actually in the languages} but with less than 1MB each.
Yup'ik and Komnzo are each part of publicly available (small) speech or text corpora~\citep{ardila-etal-2020-common,christian_dohler_2021_4695271,he-etal-2024-wav2gloss}. This means that there is a slight chance that models have encountered words in these languages before.
Short Wikipedia articles in the English Wikipedia exist for all five languages, with the ones about Yup'ik and Yélî Dnye being the most detailed, containing grammar and phonology descriptions (Iquito and Komnzo: only phonology; Sakurabiat: shorter phonology, grammar)---which means that models might recall some linguistic knowledge about these languages, if they have sufficiently memorized this data.
Iquito and Yup'ik have the largest number of resources listed in Glottolog (around 50--60)---resources that can be accessed if publicly available, but are not as likely to be included in pretraining data.
To summarize, Yup'ik is the most present in public web sources, and Iquito (P3) is the most documentationally rich but has a small general-web footprint.
Hence, LLMs are likely to be most familiar with contents in Yup'ik (P1) and Yélî Dnye (P2) (if at all) and their grammars and phonology. According to jury scores (\Cref{tab:jury_scores_problem}), Yup'ik and Sakurabiat were the best solved problems, but models struggled the most with Yélî Dnye, which indicates that \textit{web presence is not a reliable predictor of difficulty for this problem set}. 

\begin{table}[]
\centering
\begin{subtable}{0.45\textwidth}
    \centering
    \resizebox{\textwidth}{!}{%
    \begin{tabular}{lccc}
    \toprule
    Model & Language Context & Lexicon & Mean \\
    \midrule
       GPT 5.6 Sol & 90.9 & 75.0 & 83.0 \\
 Claude Opus 4.8 & 84.8 & 52.5 & 68.7 \\
\hline
 Kimi K2.6 & 93.9 & 72.5 & 83.2 \\
 Qwen 3.7 Max  & 90.9 & 72.5 & 81.7 \\
 DeepSeek V4 Pro & 93.9 & 67.5 & 80.7 \\
 Inkling & 84.8 & 60.0 & 72.4 \\
 Nemotron 3 Ultra & 90.9 & 52.5 & 71.7 \\
 MiniMax M3 & 75.8 & 42.5 & 59.1 \\
 Command A+ & 72.7 & 37.5 & 55.1 \\
 \bottomrule
    \end{tabular}
    }
    \caption{Accuracy of frontier models on 4-way MCQ knowledge probe tasks derived from IOL 2026 problems. Language Context: questions targeting language context, Lexicon: identifying real forms from the problem language of a set of four options.}
    \label{tab:knowledge_probes}
    \end{subtable}
    \hfill
  \begin{subtable}{0.45\textwidth}
  \centering
  \resizebox{\textwidth}{!}{%
  \begin{tabular}{lcc}
  \toprule
      Pairing & Pearson & Spearman \\
      \midrule
       Language Context vs GM & 0.273 & 0.195 \\
       Lexicon vs GM & 0.339 & 0.355 \\
       Mean vs GM & 0.380 & 0.430\\
       \bottomrule
  \end{tabular}
  }
  \caption{Correlating knowledge probe accuracy with overall IOL 2026 performance (automatic score, GM).}
  \end{subtable}
\caption{Results of knowledge probing tasks for a subset of benchmarked models. }
\end{table}

\textbf{Direct knowledge probes }
When we directly a probe a subset of 9 of the large benchmarked models for either contextual or lexicon knowledge about each problem language
(details in \Cref{app:probe_details}), we find that they all score fairly highly (random guess baseline would score at 25\%), as shown in \Cref{tab:knowledge_probes}. All models  score consistently higher on predicting contextual information about the languages (72--94\%) than identifying lexical items (37--75\%), indicating that they \textit{likely know more about the language than about words in the language}. Comparing their mean accuracy in probing tasks with their ranking according to automatic (GM) scores on IOL 2026, we find a positive correlation between the two (Pearson: 0.38, Spearman: 0.43). Of the two sub-tasks, correlation is higher for the lexicon probe, indicating this might be a more reliable predictor for task performance overall. This finding is strictly correlational and does not allow causal conclusions---it might be an effect of scale, generally stronger reasoning or better memorization of e.g. Wikipedia, rather than indicating any linguistics-specific training.

\begin{table}
\resizebox{\textwidth}{!}{
\begin{tabular}{lrrrrrrrr}
\toprule
 Model & chrF & EM & Geo & chrF$^{\mathrm{anon}}$ & EM$^{\mathrm{anon}}$ & Geo$^{\mathrm{anon}}$ & $\Delta$Geo & 95\% CI \\
\midrule
 GPT 5.6 Sol & 65.6 & 47.6 & 55.8 & 63.6 & 39.7 & 50.2 & -5.6 & [-16.2, +3.9]  \\
 Claude Opus 4.8 & 69.5 & 52.0 & 60.1 & 64.7 & 50.5 & 57.2 & -3.0 & [-15.2, +10.0] \\
\midrule
 Command A+ & 29.6 & 6.6 & 14.0 & 22.9 & 3.4 & 8.0 & -5.9 & [-18.0, +3.3]  \\
 Qwen 3.7 Max  & 52.8 & 27.4 & 38.0 & 42.1 & 16.5 & 26.3 & -11.7 & [-23.3, +0.6] \\
 Kimi K2.6 & 51.8 & 31.2 & 40.2 & 51.3 & 30.4 & 39.5 & -0.7 & [-16.9, +12.9]  \\
 Nemotron 3 Ultra & 36.7 & 18.0 & 25.4 & 36.8 & 19.7 & 26.7 & +1.2 & [-19.8, +18.2] \\
 DeepSeek V4 Pro & 54.2 & 28.7 & 39.4 & 57.9 & 38.4 & 46.8 & +7.4 & [-11.0, +28.2] \\
\bottomrule
\end{tabular}
}
\caption{Automatic score change when language mentions and context are dropped from the task descriptions (without custom answer parsing that would raise all scores). Scores are aggregated across three repeated runs with different random seeds.}
\label{tab:anon}
\end{table}

\begin{table}[]
    \centering
    \begin{tabular}{lrrrrr}
\hline
 Language & Geo & Geo$^{\mathrm{anon}}$ & $\Delta$Geo & 95\% CI & \# models $\downarrow$ \\
\hline
 Yélî Dnye & 21.0 & 15.2 & -5.8 & [-9.9, -1.9] & 6/7 \\
 Komnzo & 48.8 & 45.0 & -3.8 & [-10.9, +4.0] & 5/7 \\
 Sakurabiat & 42.4 & 41.4 & -1.0 & [-7.0, +4.0] & 3/7 \\
 Central Alaskan Yup'ik & 51.6 & 51.9 & +0.3 & [-9.7, +9.8] & 3/7 \\
 Iquito & 40.8 & 41.7 & +0.9 & [-3.7, +5.6] & 3/7 \\
\hline
\end{tabular}
    \caption{Automatic score change on each problem for an average across three runs from models in \Cref{tab:anon}. }
    \label{tab:anon-problem}
\end{table}

\textbf{Linguistic reasoning without language meta-information}
Each IOL task comes with a rich context and also background information about the language, e.g. language family, script, geographic region where it is spoken. Models (and humans) might leverage this knowledge to transfer from related languages they might be more familiar with. When we drop all mentions of the language and remove this context, we observe on average a loss in performance (\Cref{tab:anon}), but it is not significant according to 95\% confidence intervals with three repeated runs.
Inspecting the effect per problem language (\Cref{tab:anon-problem}), however, reveals that this removal of context is significantly hurting average model performance for Yélî Dnye (P2), with 6/7 models dropping an average of 5.8 percentage points. For this problem set, which is also the most difficult for all models, the presented language context is consistently the most helpful, because the language name itself contains a cue,\footnote{The language name ``Yélî Dnye'' constitutes a crucial cue for solving the problem 2b), item 32 (``nkéli dnye'' translating to ``English language'') as ``dnye'' means ``language''.} but overall the effect is small and noisy. 

\textbf{Qualitative inspection of reasoning traces} For models that disclose their reasoning traces (not proprietary models), we can also inspect whether they refer to typological or language family information about the problem language (given or not). A qualitative inspection reveals that language meta-information is indeed frequently present in the reasoning traces across models. This pattern is strongest for Sakurabiat problems (P4), where models refer to Tupian kinship structures and respective relational nouns, but as shown above there is no consistent effects on task scores when this association is removed. For Yélî Dnye (P2), however, models refer to priors from Papuan/Oceanic/Austronesian languages, exploring answer paths beyond the explicitly stated information (the problem states that it is a language isolate). This analysis shows that reasoning models will try to leverage prior knowledge in their reasoning, but we cannot conclude that it will help them to solve the task.

Overall, this analysis confirms that even when models have some knowledge about the problem languages or are familiar with words of these languages, \textit{it does not determine their ability to solve the respective linguistic problems}. The contextual meta-information given with each problem might bias their priors, but the difficulty, as per design of the IOL, remains in deducing patterns from the presented linguistic context, and generalizing them to new forms.

\section{Conclusion}
We introduced the IOL-AI challenge, the first evaluation of language models on genuinely unseen linguistic-olympiad problems, graded by the expert jury that grades the human contestants. Because the problems were unreleased, no model could have seen them or their solutions during training, ruling out benchmark leakage by construction. The problem languages themselves have some public documentation, but our knowledge probing shows that what models know about them does not determine success. Solving these problems requires inferring the linguistic system from the given data. This makes linguistic reasoning a true test of model reasoning capabilities.

The open challenge showed that inference-time techniques have a large impact under tight resource constraints. The winning submission more than doubled our strongest baseline with the same 14B model, with the margin coming from decoding and answer handling rather than model capacity. Prior work reported that frontier models uniformly underperform on linguistic puzzles. This is no longer the case: for the first time, models reached medal level. Claude Opus 4.8 scored above this year's gold-medal cutoff, equivalent to fourth place among 255 contestants, and Gemini 3.6 Flash reached silver. But this progress is confined to proprietary systems whose training and reasoning remain inaccessible, so it does little to advance open research on reasoning. No resource-constrained submission cleared the honorable-mention threshold, both scoring in the bottom 5\% of contestants, and even an oracle ensemble of mid-size open models lands well short of the frontier. Closing this gap with open, efficient systems is the central problem this challenge poses.

Expert grading also changes what the scores mean: it is under the jury's rubrics, not string matching, that frontier models reached medal level. Comparing the two scoring methods, automatic metrics rank systems exactly as the jury does ($\rho=1.00$, $r=0.99$), but compress the scale, inflating weak systems by ${\sim}13$ points and under-crediting strong ones, whose correct analyses in prose earn nothing automatically. We release jury evaluations, model outputs, and scores\footnote{\url{https://github.com/Cohere-Labs-Community/iol-ai-2026-results}} to support progress on both linguistic reasoning systems and their measurement.

\newpage
\section*{Limitations}

\textbf{Format Requirements}
We noticed that most models, including frontier models, struggle with the required answer format, outputting a list of individual solutions. As a consequence, many submissions to the IOL-AI challenge focused on working around format artifacts, and we had to craft post-hoc format fixing rules for frontier model outputs to make sure their output format does not perturb automatic score rankings. Future benchmarking for linguistic reasoning should re-evaluate which prompting style is most favorable for LLMs across model scales.

\textbf{Test Set Size \& Diversity} The evaluation on the IOL 2026 Individual Contest problems focuses on depth rather than breadth. As a consequence, the total number of test samples is small for testing ML models and the statistical power low. Since these tasks can be assumed to be out-of-domain for most models, the risk for high variance across multiple outputs under stochastic decoding is high, especially with long reasoning traces. Small differences in automatic scoring should therefore not be over-interpreted. The score differences between the AI solutions participating in human evaluation completely align with the jury ranking, so we can be confident that differences of that magnitude (min 5 points in GM) tend to be meaningful and human noticeable.

\textbf{Oracle Ensemble} The BoN Open ensemble uses oracle selection against the reference answers. It therefore reports an upper bound on what this pool of mid-size open models could achieve under perfect routing, rather than the performance of a deployable system. A further caveat is that a single model (Gemma4-31B-It) supplies the majority of the selected answers, so the ensemble's advantage is only partly attributable to model diversity.

\textbf{Unequal Generation Budgets} Within the mid-size tier, DeepSeek-R1-32B was run with a 32{,}768-token generation budget against 64{,}000 for the other three models (\Cref{tab:eval-hypar}), so its scores are not strictly comparable within that tier and likely understate it.

\textbf{Time Limitation} The open-science competition ran for a short time window of a month, and a longer duration might have gathered more submissions, or have allowed participants to improve their submissions further.

\textbf{Independence of Submissions} We required all submissions to be public repositories on HuggingFace, so it was possible for participants to search for each other's submission and potentially copy it. While teaming up after individual submissions was explicitly forbidden, this remained a loophole. As a consequence, we have many similar submissions, and top-scoring submissions might not credit the work of other participants that laid the foundation. Nevertheless, building on top of other peoples' work and improving upon it is also in the spirit of advancing the state of the art.

\section*{Acknowledgments}
 We thank the participants of the challenge for their contributions to advancing the understanding of linguistic reasoning challenges for LLMs. This work would not have been possible without the generous support of additional members of the IOL 2026 Jury – Mihai-Alexandru Bratu, Aida Davletova, Jan Petr, Eimear McKnight, Stanislav Gurevich – who evaluated the submissions. We especially thank Mihai, Aida and Jan for providing written impressions of the solutions. In addition, the dataset for this challenge would have not existed without the work of the greater IOL Problem Committee and the authors of this year's problems---DMM, Eimear McKnight, Lai Otsuka, and Vesko Milev. We also thank Cohere as sponsors of the compute resources for the challenge.

\bibliography{custom, anthology_cited}

\clearpage

\appendix

\section{Accessibility and Engagement}\label{app:accessibility}

To reduce barriers for participation, we hosted an online submission 101 session with a demo notebook that provides a sample submission based on a small open model. The compute provided in a T4 GPU session on Colab (session time limits are 12h for free usage) is comparable to that available in the competition, which allowed competitors to optimize model choices and decoding strategies for the time and compute limits.
In addition, we provided participants with a list of
resources about linguistic reasoning with AI, such as pointers to benchmarks and papers.

We organized this challenge intentionally outside of any academic venue to broaden participation and extend it beyond an academic audience or tie it to the requirement of paper submissions (classic shared task setup).
Instead, we advertised it heavily on social media (X, LinkedIn, Bluesky) and in open science communities.
We hosted a kick-off fireside chat with a discussion around the skills needed for linguistic reasoning both from a human perspective and the machine learning perspective.
Additionally, we reached out to authors of related works and benchmarks to invite them to participate and/or share it with their network.

\section{Technical Details}\label{app:technical}

We substantially adapted the HF competition framework for the IOL-AI competition: First, submission evaluation is re-routed from per-submission Spaces to
HF Jobs after undocumented platform rate limits. Second, every submission repository is
automatically archived for reproducibility and post-hoc human evaluation.
All modifications are publicly available in the competition Space
repository.

\section{Baselines}\label{app:baselines}

\textit{Qwen2.5-1.5B-Instruct} is the
example submission published on the competition platform and website to
show participants what is expected: a minimal script with a minimal
prompt, asking the model to answer each item on its own line.
\textit{Qwen2.5-7B-Instruct} scales this example to a larger model,
quantized to 4-bit at load time with \texttt{bitsandbytes} under the
same prompt, showing that quantization fits the compute constraints.
\textit{Qwen2.5-14B-Instruct-AWQ} illustrates the second quantization
option, shipping pre-quantized AWQ weights, to fit the T4; here the
prompt is not minimal but a reasoning prompt instructing the model on
the output format expected for each task type. Finally, \textit{Tiny
Aya Global}, a 3.35B open multilingual model, uses an
even more detailed prompt. Together, the baselines show
that larger models reach higher scores, but that careful prompt design
lets a smaller model outperform a larger one.

\Cref{app:baseline-prompts} lists the full evaluation prompts.

\section{Extended Model Benchmarking}\label{app:extended_models}
We benchmark additional strong models that are outside the resource constraints of the competition.
We prioritize models scoring highly on the Artificial Analysis Intelligence Index\footnote{\url{https://artificialanalysis.ai/evaluations/artificial-analysis-intelligence-index}} and diversity across model providers.\footnote{We do not evaluate Anthropic's Fable 5 as it requires data retention.} The evaluation hyperparameters can be found in \Cref{tab:eval-hypar}.
We leverage Linguini~\citep{sanchez2025linguini} as a development set to tune task and explanation prompt and postprocessing with an answer extraction script to further boost scores, and then use the IOL 2026 test set to select models for human evaluation.
We optimize their explanation output format for readability of the jury, to make it closer to typical IOL solutions.
\Cref{app:prompts} contain the prompts for benchmarking with and without explanations. \Cref{tab:results_extended_full}
shows the results of the automatic evaluation for the extended model list, before and after custom answer parsing.

\begin{table*}[h]
\centering
\begin{tabular}{lcccc|cccc}
\toprule
& \multicolumn{4}{c|}{Original} & \multicolumn{4}{|c}{+ custom answer parsing} \\
\textbf{Model} & \textbf{chrF} & \textbf{EM} & \textbf{GM} & \textbf{LFMR} & \textbf{chrF} & \textbf{EM} & \textbf{GM} & \textbf{LFMR} \\
\midrule
\multicolumn{5}{l|}{\textbf{Proprietary Models}} & & & & \\
\midrule
claude-opus-4-8 $\star$ & 70.6 & 52.8 & 61.0 & 92.9 & 82.9 & 68.1 & 75.1 & 100.0 \\
 gemini-3.6-flash $\star$ & 65.8 & 37.3 & 49.6 & 92.9 & 70.6 & 45.0 & 56.4 & 100.0 \\
 gpt5.6-sol  & 63.3 & 43.1 & 52.2 & 92.9 & 63.3 & 43.1 & 52.2 & 100.0 \\
\midrule
\multicolumn{5}{l|}{\textbf{Open Large Models}} & & & & \\
\midrule
kimi-k2-6 & 57.1 & 35.8 & 45.2 & 92.9 & 57.1 & 35.8 & 45.2 & 100.0 \\
 glm-5-2 & 43.5 & 21.4 & 30.5 & 71.4 & 58.0 & 35.0 & 45.0 & 92.9 \\
 deepseek-v4-pro & 49.6 & 27.5 & 36.9 & 64.3 & 53.6 & 31.5 & 41.0 & 92.9 \\
 qwen-3-7-max & 49.3 & 26.1 & 35.9 & 85.7 & 52.2 & 29.4 & 39.2 & 100.0 \\
 inkling & 45.3 & 20.6 & 30.5 & 85.7 & 48.0 & 24.3 & 34.1 & 100.0 \\
 nemotron-3-ultra & 41.5 & 20.1 & 28.9 & 64.3 & 41.5 & 20.1 & 28.9 & 64.3 \\
  command-a-plus & 33.1 & 8.3 & 16.5 & 100.0 & 34.1 & 11.1 & 19.4 & 100.0 \\
 minimax-m3 & 4.8 & 1.3 & 2.5 & 14.3 & 14.4 & 7.9 & 10.6 & 78.6 \\
 \midrule
\multicolumn{5}{l|}{\textbf{Open Mid-Size Models}} & & & & \\
  \midrule
   gemma4-31b-it   & 37.2 & 14.6 & 23.3 & 85.7 & 37.0 & 14.6 & 23.2 &  92.9 \\
   qwen3.6-27b     & 23.8 & 13.4 & 17.8 & 57.1 & 25.2 & 14.9 & 19.4 &  64.3 \\
   deepseek-r1-32b & 23.9 &  3.0 &  8.5 & 85.7 & 23.7 &  3.0 &  8.5 & 100.0 \\
   glm-4.7-flash   &  6.6 &  0.0 &  0.0 & 14.3 &  6.6 &  0.0 &  0.0 &  21.4 \\
  \hdashline
   BoN Open (oracle) $\star$ & - & - & - & - & 48.5 & 23.2 & 33.5 & 100.0 \\
\bottomrule
\end{tabular}
\caption{Evaluation scores for existing models on the IOL 2026 problem set, before and after answer parsing. The BoN Open (oracle) ensemble builds directly on custom answer parsed answers.
EM = exact match, GM = geometric mean, LFMR = line format match rate. Model outputs that were chosen for the jury are marked with $\star$.}
\label{tab:results_extended_full}
\end{table*}

\begin{table}[ht]
\centering
\small
\begin{tabular}{llccc}
\toprule
& \textbf{Language} & \textbf{chrF} & \textbf{EM} & \textbf{GM} \\
\midrule
P1 & Yup'ik      & 49.1 &  7.1 & 18.7 \\
P2 & Yélî Dnye   & 25.1 & 17.9 & 21.2 \\
P3 & Iquito      & 83.5 & 25.0 & 45.7 \\
P4 & Sakurabiat  & 32.8 & 27.8 & 30.2 \\
P5 & Komnzo      & 63.6 & 50.0 & 56.4 \\
\bottomrule
\end{tabular}
\caption{Per-problem automatic scores for the BoN Open (oracle) ensemble.
Scores are on a 0--100 scale per problem, so they are directly comparable to
jury scores expressed as a percentage of the 20 points available per problem.}
\label{tab:bon_per_problem}
\end{table}

\section{The Challenge of Format Following}\label{app:format-following}

Automatic scoring is strict about format: model answers are compared to gold by position, so a missing or shifted line can zero out a whole row even when later items are right.

\textbf{Submissions.}
Format following was a major focus in the open challenge. 34 teams used regex answer extraction, and many added per-task format prompts or retry loops. Teams also dropped chain-of-thought on scoring runs. CoT was tried by 33~teams but hurt on average, and none of the top~6 kept it: reasoning text made extraction harder and burned wall-clock time. The podium used answers-only prompts and invested in parsers instead. Because format was already enforced in the sandbox, post-hoc extraction barely moves submission scores.

\begin{figure*}[t]
  \centering
  \begin{subfigure}[t]{0.48\textwidth}
    \centering
    \includegraphics[width=\linewidth]{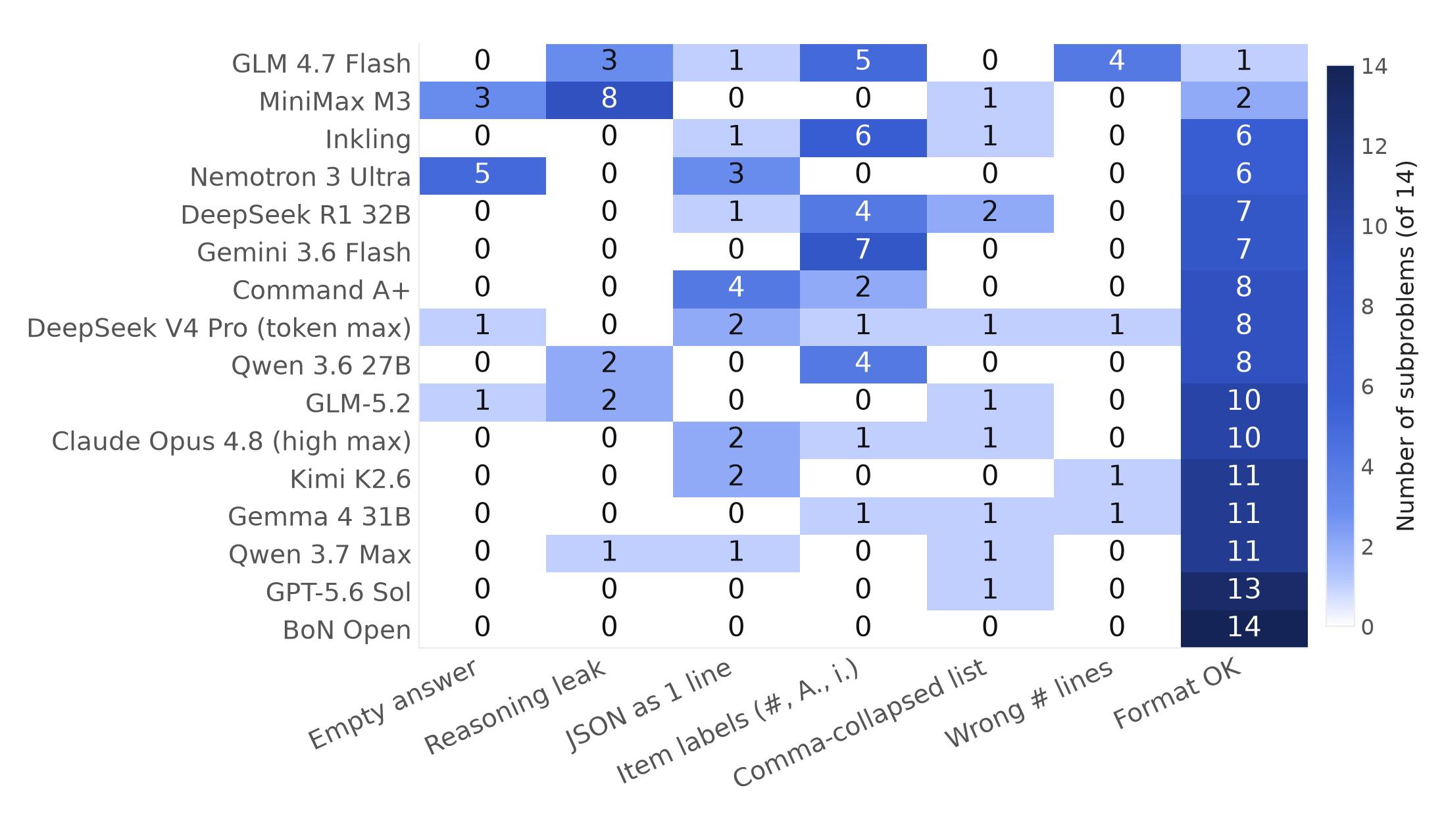}
    \caption{IOL~2026 frontier: format failures before extract.}
    \label{fig:format-iol-before}
  \end{subfigure}
  \hfill
  \begin{subfigure}[t]{0.48\textwidth}
    \centering
    \includegraphics[width=\linewidth]{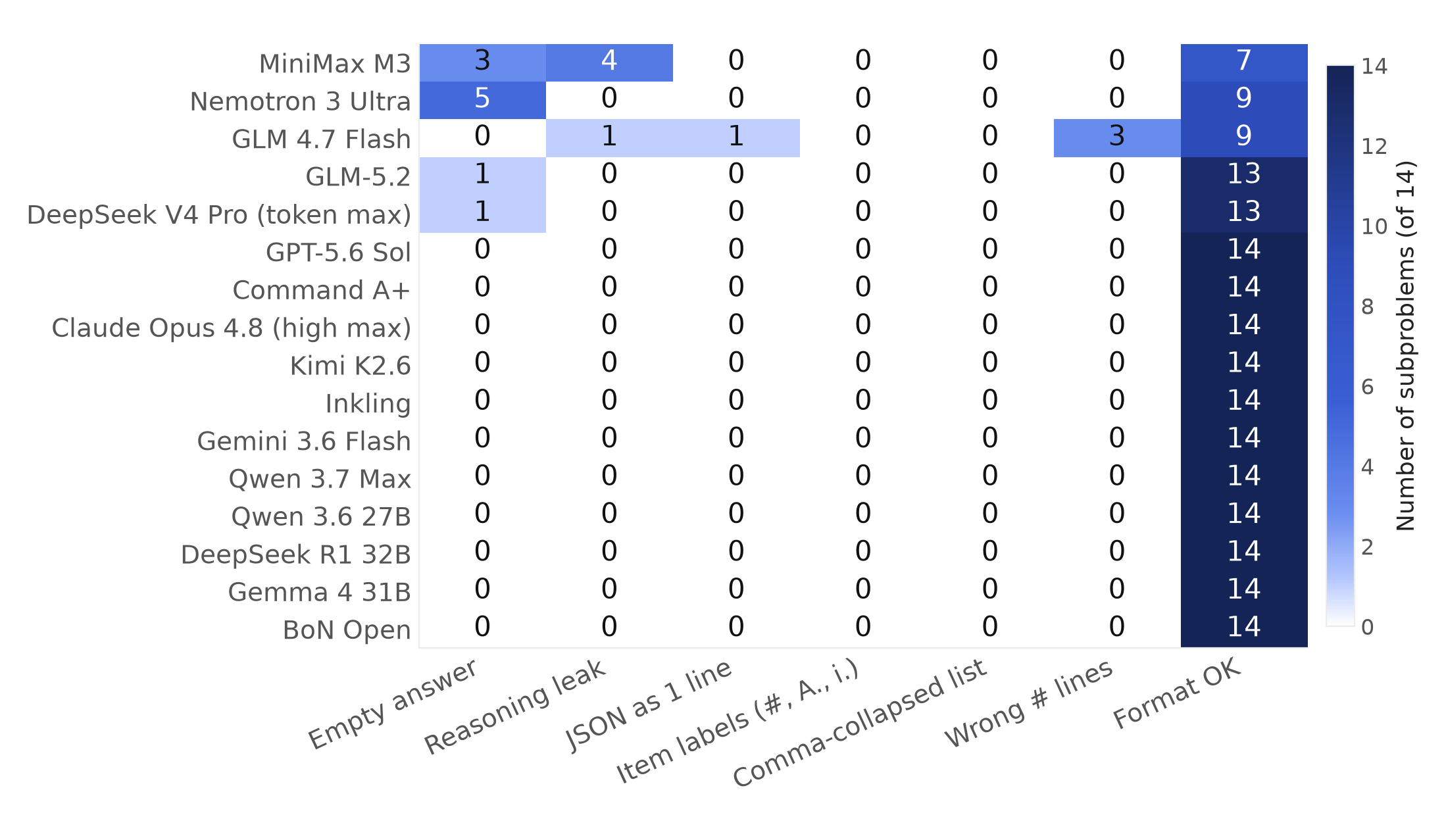}
    \caption{IOL~2026 frontier: format failures after extract.}
    \label{fig:format-iol-after}
  \end{subfigure}
  \caption{Format-failure prevalence for IOL~2026 frontier runs
  before and after light answer extraction.}
  \label{fig:format-iol}
\end{figure*}

\textbf{Frontier models.}
Format errors are still common in raw frontier explain outputs. Light postprocessing removes most of them (\Cref{fig:format-iol}): empty answers cannot be fixed, but superficial non-conformities such as included item labels or JSON format as a one-liner are easily repaired. \Cref{tab:results_extended_full} reports scores before and after custom answer parsing. Parsing never lowers the score. At best it raises GM by about 14 points (e.g.\ GLM-5.2 from 30.5 to 45.0; Claude Opus from 61.0 to 75.1). Already clean runs stay unchanged.

\FloatBarrier
\section{Evaluation Prompts}\label{app:prompts}

\subsection{Benchmarking Task Prompts}\label{app:benchmark-prompts}
For Linguini evaluations, we use the task prompt given in \Cref{fig:linguini_prompt}.

\begin{figure*}[!htb]
\begin{lstlisting}[breaklines=true, basicstyle=\ttfamily\small,
columns=fixed, keepspaces=true, showstringspaces=false,
xleftmargin=0pt, gobble=0]
You are solving a linguistic puzzle. All the information you need is contained in the context below - no prior knowledge of this language is required.

Context:
[CONTEXT]

Question:
[QUESTION]

Answer with only the requested word or phrase. Do not include explanations in your final answer.
\end{lstlisting}
\caption{Linguini evaluation prompt.}\label{fig:linguini_prompt}
\end{figure*}

For out-of-the-box benchmarking of frontier/open models where explanations are required for human evaluation, we use the task prompt given in \Cref{fig:task_prompt}. We verified on Linguini that the added explanation requirement does not have substantial impact on task performance of the top-scoring models participating in the human evaluation.

\begin{figure*}[!htb]
\begin{lstlisting}[breaklines=true, basicstyle=\ttfamily\small,
columns=fixed, keepspaces=true, showstringspaces=false,
xleftmargin=0pt, gobble=0]
You are solving a linguistic puzzle. All the information you need is contained in the context below - no prior knowledge of this language is required.

Context:
[CONTEXT]
Question:
[QUESTION]

Give your final answer, then an explanation of how you arrived at it.

In the final answer, only include the requested word(s) or phrase(s), without markdown.
In the explanation, summarize your linguistic analysis, and use tables or schemata (in markdown) to illustrate derived rules.
Do not write a long chain-of-thought or lengthy narration.

Use this format exactly:
Answer:
<your final answer only - the requested word(s) or phrase(s)>

Explanation:
<your explanation>
\end{lstlisting}
\caption{Task prompt for Linguini benchmarking of out-of-the-box models, including explanations.}\label{fig:task_prompt}
\end{figure*}

For IOL we slightly modified it because we also allowed for answers in JSON format, see \Cref{fig:prompt_final}.

\begin{figure*}[!htb]
\begin{lstlisting}[breaklines=true, basicstyle=\ttfamily\small,
columns=fixed, keepspaces=true, showstringspaces=false,
xleftmargin=0pt, gobble=0]
You are solving an International Linguistics Olympiad problem. All the information you need is contained in the context below - no prior knowledge of this language is required.

Context:
[CONTEXT]
Question:
[QUESTION]

Answer every numbered (or lettered) item in the question, in order.
Return either a JSON list of strings, or one answer per line (no numbering).

Give your final answer, then an explanation of how you arrived at it.

In the final answer, only include the requested word(s) or phrase(s), without markdown.
In the explanation, summarize your linguistic analysis, and use tables or schemata (in markdown) to illustrate derived rules.
Do not write a long chain-of-thought or lengthy narration.

Use this format exactly:
Answer:
<your final answer only - the requested word(s) or phrase(s)>

Explanation:
<your explanation>
\end{lstlisting}
\caption{Final IOL task prompt for benchmarking out-of-the-box models, including explanations.}\label{fig:prompt_final}
\end{figure*}

\subsection{Baseline System Prompts}\label{app:baseline-prompts}
The four organizer baselines (\Cref{tab:baselines}) use a chat format with a
separate system prompt, unlike the single task prompts above. We list the system
prompt and the user template for each; \texttt{\{context\}} and \texttt{\{query\}}
are filled in per test item.

\begin{figure*}[!htb]
\begin{lstlisting}[breaklines=true, basicstyle=\ttfamily\small,
columns=fixed, keepspaces=true, showstringspaces=false,
xleftmargin=0pt, gobble=0]
SYSTEM: You solve International Linguistics Olympiad problems. Answer every numbered item. Put each answer on its own line, in order, with no numbering and no extra text.

USER:
{context}
{query}
\end{lstlisting}
\caption{Qwen2.5-1.5B-Instruct (starter) and Qwen2.5-7B-Instruct (NF4) baseline prompt.}
\end{figure*}

\begin{figure*}[!htb]
\begin{lstlisting}[breaklines=true, basicstyle=\ttfamily\small,
columns=fixed, keepspaces=true, showstringspaces=false,
xleftmargin=0pt, gobble=0]
SYSTEM:
You solve International Linguistics Olympiad problems by reasoning from the data you are given. You may meet a task type you have never seen: read the instruction and the examples, and answer in the same form they use. Common task types and what to give -- translation: the translated form only, in the language the task asks for; fill_blanks: only the missing form for each blank; match_letters: only the option letter (for example A, B, C); text_to_num: the number in digits; num_to_text: the number written out in words, in the language asked; any other type: give exactly what the instruction asks, nothing else. Reason step by step first. Then write a line that says exactly FINAL ANSWERS: and, below it, one answer per line in the order the items are asked -- the bare answer only, no numbering, no quotes, no extra text.

USER:
{context}
{query}
\end{lstlisting}
\caption{Qwen2.5-14B-Instruct-AWQ baseline prompt}
\end{figure*}

\begin{figure*}[!htb]
\begin{lstlisting}[breaklines=true, basicstyle=\ttfamily\small,
columns=fixed, keepspaces=true, showstringspaces=false,
xleftmargin=0pt, gobble=0]
SYSTEM:
You solve International Linguistics Olympiad problems by reasoning from the data in CONTEXT you are given to solve the problems in QUERY.
There are common TASK TYPES that we specify below, but you may meet a TASK TYPE you have never seen: read the instruction and the examples, and answer the QUERY in the same form they use.
Common TASK TYPES and what to return:
`translation`: return the translated form only, in the language the task asks for;
`fill_blanks`: return only the missing form for each indicated blank (beware: this could be many different things: a word, a part of a word or a phonetic transcription---pay close attention to what part of the CONTEXT is missing in QUERY);
`match_letters`: return only the option letter (for example A, B, C);
`text_to_num`: return the number in digits;
`num_to_text`: return the number written out in words, in the language asked;
any other type: return exactly what the instruction asks for, nothing else.
As the first part of your answer, reason step by step about (1) the linguistic rules that can be deduced from the given examples in CONTEXT, and (2) how to apply them to the given problems in QUERY, and (3) in what format answers need to be returned (words, numbers, phonetic transcriptions, ...).
Then write a draft of the final answer. Subsequently, compare it with the format requirements again, and verify it's compliant with the deduced rules, and it is complete, i.e. has an answer for each element in QUERY. If necessary, correct and refine. Finally, write a line that says exactly `FINAL ANSWERS:` and, below it, write the answers to the items requested in QUERY (not those in CONTEXT), one answer per line (separated by newline) in the order the items are asked for in the QUERY -- the bare answer only, no numbering, no quotes, no extra text, according to the given TASK TYPE.

USER:
CONTEXT:{context}
TASK TYPE:`{task_type}`
QUERY:{query}
\end{lstlisting}
\caption{Tiny Aya Global baseline prompt}
\end{figure*}

\section{Grading of AI Solutions by the IOL Jury}\label{app:jury}

In contrast to the in-person grading at IOL, due to time constraints at the event, grading of the AI solutions is done remotely in the week following IOL. 1--3 jury members from each grading group (that graded each of the 5 problems at IOL) volunteered to grade the AI solutions for each problem. Each jury member received pdf printouts of the AI solutions and explanations, anonymized with respect to model information.

Using the same grading schemes that were used at IOL (which were crafted by each grading group in the months leading to the event), each jury member then graded each AI paper individually, assigning points for each criterion specified in the grading scheme in a large spreadsheet. The spreadsheet is designed so that individual jurors can work separately in their own sheets, and afterwards a comparison sheet highlights any \emph{discrepancies} between the grades of 2 or more jurors. At IOL, each contestant's paper is graded by at least 2 jurors and discrepancies are resolved largely by in-person discussion. For the challenge, all problems were graded twice except problem 3, which was graded 3 times. For problem 4, the IOL jury chair
served as a second grader despite not being part of that grading group. Discrepancies were resolved largely by virtual discussion (through text), and in a couple of cases by decision of the jury chair.

Once all discrepancies were resolved, the scores for the 5 problems were summed up into one total score (out of 100) per submission.

\section{Submissions}\label{app:submissions}

\textbf{Submission Statistics}
Participants could form teams of up to three, but most competed
individually: only four teams of two were formed. Submissions were
back-loaded, with 43\% arriving in the final 48 hours.
590 (80.7\%) submissions ran successfully and received a score. The 141 failures
fall into four families: library and environment mismatches between the
submission repository and the evaluation image (54, 38.3\%), errors in
the participant inference script or at runtime (32, 22.7\%), exceeding
the wall-clock or memory limits (27, 19.1\%), and offline packaging
mistakes such as code still attempting to reach the Hub at evaluation
time (19, 13.5\%); the remaining 9 (6.4\%) fall outside these families.
Common fixes included shipping model weights inside the repository,
bundling updated wheels for offline install, quantizing to fit the T4
(AWQ, NF4, or GGUF), and, against the timeout, using adaptive token
budgets and writing \texttt{submission.csv} incrementally so a killed
run still scores.

\textbf{Leading Pipelines}
The seven highest-ranked entries all reuse our minimal system prompt
verbatim (\Cref{app:baseline-prompts}), and all lower the repetition
penalty except Hul (6), which keeps the default. The entries that sample
draw up to 8 retries (ranks 1, 4, 5) or up to 24 (ranks 2, 3) at
temperature 0.5, for as long as time allows, and a retry only overrides
the greedy answer when several samples agree on the same alternative;
friedspaghetti (4) counts near-identical strings as the same answer, so
formatting variants do not split the vote. Hul and jbuaba (7) never
sample. Token budgets are either derived from the time remaining (ranks
1, 3, 4, 5, between 192 and 900 tokens per problem) or fixed at 512
tokens (ranks 2, 6, 7). Incremental writing means a complete submission
file is on disk before the model even loads, and is rewritten after
every pass or row, so a crash or timeout still leaves a valid file.
Post-processing is where the entries differ most: most pad or trim the
answers to the expected number of items, Hul normalises them by task
type, and jbuaba answers matching problems by reading the model's
probability for each option letter and assigning options so that none
repeats.

\textbf{Submission Model Use}
Nine of the top ten entries run a 14B Qwen model: eight use
Qwen2.5-14B-Instruct-AWQ and one uses Qwen3-14B; the remaining entry, at
rank 9, runs a 4.2B Qwen3.5 model in 8-bit GGUF. Across the 42 teams with
a scored final submission, model choice is similarly concentrated: 29
teams selected Qwen2.5-14B-Instruct-AWQ for their final
submissions.\footnote{We identified each backbone from the weights
shipped at the scored commit, reading \texttt{config.json} or the GGUF
header rather than repository or file names.} This concentration has two
causes: Qwen2.5-14B-Instruct-AWQ is the largest model that fits on the T4
within the time limit, and it was the model presented in our submission
101 session. Most teams therefore adopted it as a starting point and
worked on the pipeline around it. The best submission more than doubles
our strongest baseline with the same model (GM 9.40 $\rightarrow$ 19.79),
so the margin at the top comes from decoding and output handling rather
than model capacity.

\section{Evaluation Hyper-parameters}
\label{app:Eval-Hypar}

\Cref{tab:eval-hypar} reports the evaluation configurations that were used for benchmarking frontier and open models. We set temperature and top $p$ to whatever was the recommended configuration in model cards and the reasoning effort and token limit as high as possible.

\begin{table*}[h]
\centering
\begin{tabular}{lccccc}
\toprule
\textbf{Model} & \textbf{Provider} & \textbf{temp} & \textbf{top\_p} & \textbf{Reasoning Effort} & \textbf{Max Tokens}\\
\midrule
Claude Opus 4.8 & Bedrock  & 1.0 & 1.0 & high & 128000 \\
GPT-5.6 & OpenAI & 1.0 & 1 & high & 128000 \\
Kimi K2.6 & TogetherAI & 1.0 & 0.95 & N/A & 256000 \\
GLM-5.2 & TogetherAI & 1.0 & 0.95 & max & 128000 \\
DeepSeek V4 Pro & TogetherAI & 1.0 & 1 & high & 128000 \\
Qwen3.7-Max & TogetherAI & 0.6 & 0.95 & N/A & 65536\\
Inkling & TogetherAI & 1.0 & 0.95 & high & 256000 \\
Nemotron 3 Ultra & TogetherAI & 0.3 & 0.95 & high & 65536 \\
Command A Plus & Cohere & 0.6 & 0.95 & N/A & 64000 \\
MiniMax M3 & TogetherAI & 1.0 & 0.95 & N/A & 131072 \\
Gemini 3.6 Flash & Google API & N/A & N/A & high & 65536 \\
\midrule
Gemma 4 31B & Self-hosted & 1.0 & 0.95 & N/A & 64000 \\
Qwen3.6 27B & Self-hosted & 0.6 & 0.95 ($k$=20) & N/A & 64000 \\
DeepSeek R1 32B & Self-hosted & 0.6 & 0.95 & N/A & 32768 \\
GLM 4.7 Flash & Self-hosted & 0.6 & 0.95 ($k$=40)$^{\dagger}$ & N/A & 64000 \\
\bottomrule
\end{tabular}
\caption{Hyper-parameters for evaluation. Self-hosted models were served with vLLM; ``Max Tokens'' is the generation budget, not the model's context window.
$^{\dagger}$GLM 4.7 Flash additionally requires \texttt{repetition\_penalty}${=}1.05$; without it the model degenerates into repetition loops, which is reflected in its low format-compliance and near-zero scores.}
\label{tab:eval-hypar}
\end{table*}

\section{Linguini Scores}
\Cref{tab:results_linguini} contains automatic evaluation results for the Linguini benchmark \citep{sanchez2025linguini}.

\begin{table*}[h]
\centering
\begin{tabular}{l|cccc|cccc}
\toprule
& \multicolumn{4}{c|}{Original} & \multicolumn{4}{c}{+ custom answer parsing} \\
\textbf{Model} & \textbf{chrF} & \textbf{EM} & \textbf{GM} & \textbf{LFMR} & \textbf{chrF} & \textbf{EM} & \textbf{GM} & \textbf{LFMR} \\
\midrule
\multicolumn{9}{l}{\textbf{Closed Models}} \\
\midrule
 GPT 5.6 Sol & 72.0 & 26.9 & 44.0 & 90.6 & 78.5 & 33.8 & 51.5 & 98.8 \\
 Claude Opus 4.8 (Bedrock) & 54.8 & 6.9 & 19.4 & 81.9 & 71.1 & 26.9 & 43.7 & 94.4 \\
 Gemini 3.6 Flash & 63.7 & 10.6 & 26.0 & 99.4 & 78.0 & 33.1 & 50.8 & 98.8 \\
\midrule
\multicolumn{9}{l}{\textbf{Open Large Models}} \\
\hline
 Qwen 3.7 Max & 58.2 & 12.5 & 27.0 & 91.2 & 65.3 & 19.4 & 35.6 & 96.3 \\
 DeepSeek V4 Pro & 48.9 & 10.0 & 22.1 & 66.9 & 55.7 & 19.4 & 32.9 & 74.4 \\
 Kimi K2.6  & 40.1 & 11.2 & 21.2 & 51.2 & 44.4 & 18.8 & 28.8 & 57.5 \\
 GLM 5.2 & 47.3 & 8.1 & 19.6 & 66.9 & 55.9 & 16.9 & 30.7 & 79.4 \\
 Nemotron 3 Ultra & 39.1 & 8.1 & 17.8 & 50.6 & 40.3 & 9.4 & 19.4 & 61.9 \\
 Inkling & 49.8 & 3.8 & 13.7 & 57.5 & 59.4 & 13.8 & 28.6 & 95.0 \\
 Command A+  & 40.3 & 1.9 & 8.7 & 64.4 & 41.6 & 2.5 & 10.2 & 85.6 \\
 MiniMax M3 & 17.7 & 3.1 & 7.4 & 24.4 & 24.9 & 5.0 & 11.2 & 41.9 \\
 \midrule
\multicolumn{9}{l}{\textbf{Open Mid-Size Models}} \\
  \midrule
  Qwen 3.8 27B (think)      &      50.4  &  9.4 &  21.7  & 73.1 & 53.7  & 11.9 &
  25.2  & 88.1\\
  Muse Glimmer 30B (high)  &      49.9   & 8.8 &  20.9  & 91.2 & 52.3   &10.0&
  22.9  & 92.5\\
   Gemma 4 31B   & 45.4 & 3.1 & 11.9 & 93.8 & 50.7 & 6.2 & 17.8 & 93.8  \\
   
   Qwen 3.6 27B     & 44.0 & 1.9 &  9.1 & 83.1 & 50.0 & 5.0 & 15.8 & 95.0 \\
   Qwen 3.6 27B (think)      &      41.1   & 2.5  & 10.1 &  79.4 & 45.1  &  5.6&
  15.9 &  87.5\\
    DeepSeek R1 32B & 29.1 & 0.0 &  0.0 & 83.8 & 32.7 & 0.6 &  4.5 & 85.0 \\
   GLM 4.7 Flash   &  5.6 & 0.6 &  1.9 & 18.1 &  7.4 & 0.6 &  2.1 & 21.2 \\
  \hdashline
   BoN Open (oracle) & 50.5 & 4.4 & 14.9 & 90.0 & 57.5 & 8.8 & 22.4 & 100.0 \\
\bottomrule
\end{tabular}
\caption{Out-of-the-box benchmarking scores for existing models on Linguini. EM = exact match, GM = geometric mean, LFMR = line format match rate.}
\label{tab:results_linguini}
\end{table*}


\section{Knowledge Probing Details}\label{app:probe_details}
In order to turn the language context per problem into probing tasks, we analyze each problem with an LLM, and prompt it to extract the relevant information (either contextual information, or lexical items from the problem language). \Cref{tab:probe_examples} shows examples. All tasks and answer options are verified by a human annotator who edited especially alternative response options to avoid making the right answer option too obvious. In total they span 33 questions about language context, and 40 about lexical items.

\begin{table}[t]
    \centering
   \resizebox{\columnwidth}{!}{%
    \begin{tabular}{lllp{6.5cm}}
        \toprule
        Task & Problem Language & Question & Answer Options \\
        \midrule
        \multirow{5}{*}{\shortstack[l]{Language\\Context}}
          & Yup'ik
            & Region where it is spoken?
            & A.~General Central Yup'ik; B.~eastern Alaska; C.~northern Alaska; \textbf{D.~southwestern Alaska} \\
        
          & Yélî Dnye
            & Island / place where it is spoken?
            & A.~Brazil; B.~New Britain; \textbf{C.~Rossel Island}; D.~Norton Sound \\
        
          & Iquito
            & Does the language have tones?
            & A.~not known; \textbf{B.~yes}; C.~no; D.~in some local dialects \\
        
          & Sakurabiat
            & Approximate number of speakers?
            & A.~3; B.~300000; \textbf{C.~13}; D.~13000 \\
        
          & Komnzo
            & Village where it is spoken?
            & A.~Tupian; B.~everywhere in south-eastern Papua New Guinea; \textbf{C.~Rouku}; D.~Jipai \\
        \midrule
        \multirow{5}{*}{Lexicon}
          & Yup'ik
            & Real word/phrase?
            & A.~at'aiqnt; B.~aknaepvn; C.~oykey; \textbf{D.~nanevpak} \\
        
          & Yélî Dnye
            & Real word/phrase?
            & A.~iikkinin; B.~otak; \textbf{C.~kinikini}; D.~hiîtnêg: \\
        
          & Iquito
            & Real word/phrase?
            & A.~nat'rarkaq; \textbf{B.~paapaaja}; C.~niieeani; D.~paaaaajp \\
        
          & Sakurabiat
            & Real word/phrase?
            & \textbf{A.~abitop}; B.~okokty; C.~ibptoa; D.~nat'rarkaq \\
        
          & Komnzo
            & Real word/phrase?
            & A.~zsmawmk; B.~shatäzw; C.~tep'liuq; \textbf{D.~zwäsath} \\
        \bottomrule
    \end{tabular}%
    }
    \caption{Examples of knowledge probing tasks derived from the problem statements from IOL 2026. Bold-faced answers are the correct answers. Language Context = 4-way MCQ over meta-facts from the IOL blurbs; Lexicon = 4-way MCQ over attested forms (distractors: scrambled or other-language). Questions are shortened.}
    \label{tab:probe_examples}
\end{table}

\end{document}